\documentclass[conference]{IEEEtran}
\IEEEoverridecommandlockouts

\usepackage{cite}

\usepackage{amsmath,amssymb,amsfonts}
\usepackage{algorithmic}
\usepackage{graphicx}
\usepackage[inkscapelatex=false]{svg}
\usepackage{textcomp}
\usepackage{xcolor}
\usepackage{graphicx}
\usepackage{float}
\usepackage{dblfloatfix}
\usepackage{subcaption} 
\usepackage{wrapfig} 
\usepackage{tcolorbox}
\usepackage[dvipsnames]{xcolor}
\usepackage{placeins}
\usepackage{booktabs}
\usepackage{multirow}
\usepackage{balance}
\usepackage{hyperref}

\newcommand{\halfcolbox}[1]{
    \begin{center}
        \begin{tcolorbox}[
            width=1.0\columnwidth,       
            arc=2mm,                     
            auto outer arc,
            boxrule=1.0pt,               
            colback=Apricot,               
            colframe=black,              
            halign=center,               
            valign=center,               
            fontupper=\itshape,          
            boxsep=2pt,                  
            ]
            #1
        \end{tcolorbox}
    \end{center}
}
\def\BibTeX{{\rm B\kern-.05em{\sc i\kern-.025em b}\kern-.08em
    T\kern-.1667em\lower.7ex\hbox{E}\kern-.125emX}}

\begin{document}

\title{Latency Analysis and Optimization of Alpamayo 1 via Efficient Trajectory Generation}




\author{
\IEEEauthorblockN{
Yunseong Jeon\IEEEauthorrefmark{1},
Namcheol Lee\IEEEauthorrefmark{2},
Yoonsu Lee\IEEEauthorrefmark{1},
Jangwoon Park\IEEEauthorrefmark{2},
Sol Ahn\IEEEauthorrefmark{1},
Jong-Chan Kim\IEEEauthorrefmark{1},
Seongsoo Hong\IEEEauthorrefmark{2}
}

\IEEEauthorblockA{\IEEEauthorrefmark{1}
Graduate School of Automobile and Mobility, 
Kookmin University, Seoul, Republic of Korea}

\IEEEauthorblockA{\IEEEauthorrefmark{2}
Department of Electrical and Computer Engineering, 
Seoul National University, Seoul, Republic of Korea}

\IEEEauthorblockA{
\{kb0316, dldbs0319, solahn00, jongchank\}@kookmin.ac.kr,
\{nclee, jwpark, sshong\}@redwood.snu.ac.kr
}
}

\maketitle

\begin{abstract}
Reasoning-based end-to-end (E2E) autonomous driving has recently emerged as a promising approach to improving the interpretability of driving decisions as it can generate human-readable reasoning together with predicted trajectories. Such approaches commonly generate multiple trajectories to capture diverse future behaviors, and they fall into two categories: (1) multi-reasoning, where one reasoning sequence is generated per trajectory, and (2) single-reasoning, where a single reasoning is shared across all trajectories. The former offers richer diversity at the cost of redundant computation, while the latter is more efficient but is often assumed to sacrifice diversity. Alpamayo 1, a representative system, adopts the multi-reasoning approach and achieves competitive trajectory prediction performance. However, the efficiency of this design remains largely unexplored, making it a well-motivated subject for investigation. In this paper, we systematically analyze and improve Alpamayo 1 in two ways. First, we reduce inference latency while preserving trajectory diversity by redesigning Alpamayo 1 into a single-reasoning system. Through extensive experiments, we find that replacing multi-reasoning with single-reasoning does not meaningfully degrade trajectory diversity. Second, we accelerate diffusion-based action generation by eliminating inter-block overhead arising from unnecessary copy operations and inefficient kernel execution. Through closed-loop and open-loop experiments, we validate both optimizations, demonstrating a 69.23\% reduction in inference latency while maintaining trajectory diversity and prediction quality. These results highlight the importance of jointly analyzing system architecture and runtime execution to improve the efficiency of reasoning-based E2E AD systems.

\end{abstract}

\begin{figure*}[!t]
    \centering
    \includegraphics[width=\linewidth]{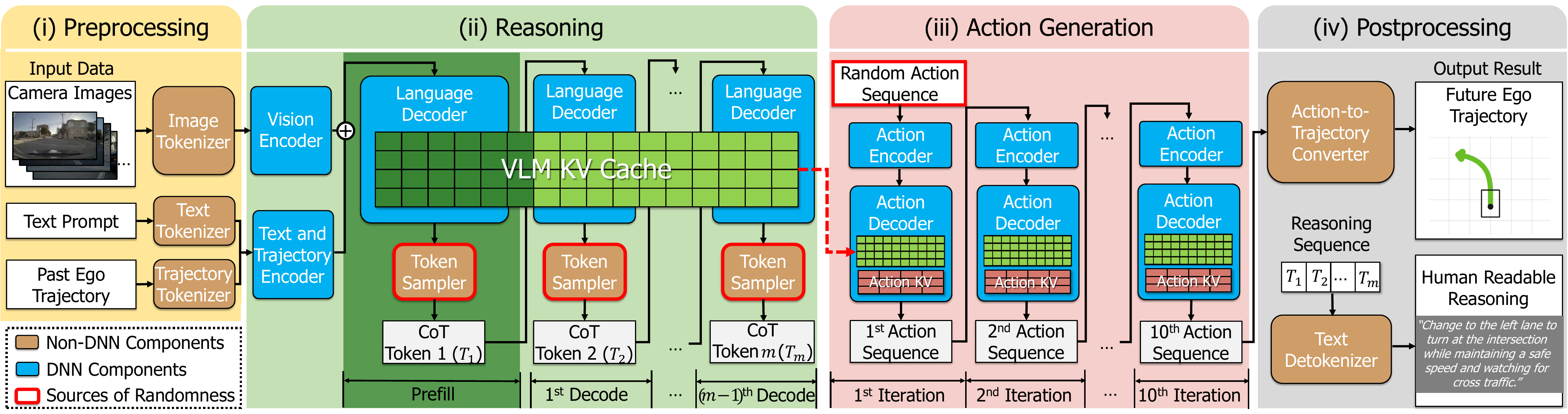}
    \caption{Runtime Architecture of Alpamayo 1 (hereinafter Alpamayo)}
    \label{fig:architecture}
\end{figure*}

\section{Introduction}

End-to-end (E2E) autonomous driving has recently gained significant attention as a paradigm that directly maps sensor observations to driving actions or trajectories using deep neural networks (DNNs)~\cite{codevilla2018end, VAD, chitta2022transfuser, DriveTransformer, UniAD, diffusiondrive}. While this approach simplifies the traditional modular approaches, it often suffers from limited interpretability, making it difficult to understand the rationale behind driving decisions. To address this limitation, {\em reasoning-based} approaches have recently emerged as a promising direction~\cite{Alpamayo, EMMA, DriveGPT4, GPTDriver, DriveVLM, Senna, AutoVLA, AutoDriveR2, AdaThinkDrive, FutureSightDrive, ReasonPlan, ORION, DiffVLA, IRL-VLA, ColaVLA}. These models augment trajectory generation with large language model (LLM)-based reasoning, enabling better transparency and diagnosability.

Since driving scenarios often admit multiple plausible futures, reasoning-based models commonly generate multiple candidate trajectories to capture diverse future driving behaviors. In this multi-trajectory generation, the relationship between reasoning generation and trajectory generation gives rise to two distinct design paradigms. The first is the {\em multi-reasoning} approach, where a separate reasoning is generated for each predicted trajectory~\cite{Alpamayo, EMMA, DriveGPT4, GPTDriver, DriveVLM, Senna, AutoVLA, AutoDriveR2, AdaThinkDrive, FutureSightDrive, ReasonPlan}. This design allows each trajectory to be conditioned on its own reasoning, potentially improving behavioral diversity. However, it also introduces significant computational redundancy because reasoning has to be repeatedly generated for each trajectory. The second paradigm is the {\em single-reasoning} approach, in which a single reasoning is shared across all predicted trajectories~\cite{ORION, DiffVLA, IRL-VLA, ColaVLA}. While this design is computationally more efficient, it is commonly assumed to reduce trajectory diversity because all trajectories rely on the same reasoning. Nonetheless, since trajectory generation is inherently stochastic, the single-reasoning approach can still capture diverse behaviors.

Among recent reasoning-based autonomous driving models, {\em Alpamayo 1} from NVIDIA (hereinafter Alpamayo) is a representative vision-language-action (VLA) model that integrates vision-language model (VLM)-based reasoning with diffusion-based action (i.e., steering and acceleration) generation (actions can be converted into trajectories) and achieves competitive trajectory prediction performance~\cite{Alpamayo}. Despite its effectiveness, the efficiency of this model and its inference architecture remains largely unexplored. In particular, Alpamayo's multi-reasoning design together with iterative diffusion-based action generation introduces substantial latency overhead as the number of trajectories grows, which can limit its practicality in real-time driving systems. This motivates a careful analysis of the architectural and runtime factors that contribute to the latency of Alpamayo.\par

In this study, we perform a comprehensive analysis of Alpamayo's runtime architecture and propose complementary optimizations to significantly reduce its inference latency while preserving trajectory diversity and prediction quality. We first conduct a detailed architectural dissection of Alpamayo, examining each component and its role in the inference pipeline. Based on this dissection, we perform a fine-grained latency analysis that reveals the following two dominant bottlenecks. First, the multi-reasoning design requires a separate reasoning to be generated for each trajectory, resulting in inference latency that scales linearly with the number of trajectories. Second, the diffusion-based action generation incurs substantial latency overhead during denoising steps. It is caused by unnecessary memory copies and reallocations for the dynamic KV cache management and by the repeated fine-granular GPU kernel launches from CPU that prevents continuous GPU kernel executions.

Guided by these findings, we propose two targeted optimizations. First, we redesign Alpamayo into a single-reasoning architecture, where a single reasoning conditions all trajectory generation, eliminating redundant computation. Through extensive experiments, we demonstrate that this redesign does not meaningfully degrade trajectory diversity and prediction quality, challenging the widely held assumption that multi-reasoning is necessary for diverse predictions. Second, we eliminate the identified latency overhead in the action generation. For that, we implement a static KV cache management scheme that preallocates a static KV buffer and reuses it throughout the diffusion iterations. This approach is feasible because we assume a static periodic application (i.e., autonomous driving), where its maximum KV cache size can be precisely estimated by profiling the KV cache growth pattern offline. Also, we apply the CUDA graph capture technique to minimize the GPU kernel launch overhead by recording a sequence of GPU kernel launch commands and just replaying it without additional launch overhead. This can be possible since we are using a static KV cache with a known KV buffer address that does not change throughout the action generation process.


We evaluate the proposed optimizations through open-loop experiments on the NVIDIA Physical AI Dataset~\cite{PhysicalAIDataset} and closed-loop experiments on the AlpaSim autonomous driving simulator~\cite{AlpaSim}. The results demonstrate up to a 69.23\% reduction in inference latency while maintaining comparable trajectory diversity and prediction quality.

The contributions of this study can be summarized as:
\begin{itemize}
    \item We present a detailed breakdown of Alpamayo's inference architecture and a per-component latency analysis obtained by profiling its runtime execution.
    \item We compare the single-reasoning and multi-reasoning approaches for multi-trajectory generation and redesign Alpamayo into a single-reasoning system.
    \item We reduce the inference latency by removing redundant memory copy overhead and kernel launch delays during diffusion-based action generation.
\end{itemize}

The remainder of this paper is organized as follows. Section~\ref{sec:alpamayo_overview} presents an overview of Alpamayo. Section~\ref{sec:delay component} conducts a detailed latency analysis, redesigns Alpamayo into a single-reasoning architecture to resolve the primary bottleneck, and identifies diffusion-based action generation as the remaining bottleneck. Section~\ref{sec:optimizing} presents runtime optimizations for this bottleneck. Section~\ref{sec:experiments} reports experimental results, Section~\ref{related} reviews related work, and Section~\ref{conclusion} concludes the paper.\par

\section{Alpamayo Overview}
\label{sec:alpamayo_overview}

\subsection{Overall Architecture}

The official description of Alpamayo can be found in~\cite{Alpamayo}. However, it focuses only on the neural architecture and the training method of the model itself without describing the runtime inference architecture. In this regard, we analyze the runtime architecture of Alpamayo based on the open source inference code officially released by NVIDIA, which comprises 3,462 lines of Python source code~\cite{alpamayo_github}. The codebase is based on PyTorch~\cite{Pytorch2019} and the Hugging Face Transformers library~\cite{HuggingFace-Transformers}.

Alpamayo is a VLA model composed of several DNN models, among which the following three transformer-based models are the most central:
\begin{itemize}
\item {{\bf Vision encoder} converts input image tokens into visual token embeddings, which are multi-dimensional vector representations of image tokens that can be understood by language models in addition to text prompts (composed of 27 transformer blocks).}
\item {{\bf Language decoder} does the actual language-based reasoning to produce so called the chain-of-thought (CoT) tokens such that the reasoning can be understood by humans while the reasoning is correctly reflected in the trajectory generation afterwards (36 transformer blocks).}
\item {{\bf Action decoder} generates a sequence of actions (i.e., steering and acceleration) to be executed by the ego vehicle for the next time steps, which eventually can be interpreted as a 2-dimensional trajectory for visualization and evaluation (36 transformer blocks).}
\end{itemize}
In general, an LLM is technically a language decoder that produces output words (i.e., tokens) in an autoregressive manner from given input prompts. An LLM together with  a vision encoder is called a VLM that can interpret images in addition to languages. If a VLM is working with an action decoder in behind, just like Alpamayo, they are collectively recognized as a VLA model.

Fig.~\ref{fig:architecture} depicts the runtime inference architecture of Alpamayo, which is composed of four modules: (i) preprocessing, (ii) reasoning, (iii) action generation, and (iv) postprocessing. In each module, the blue components are DNNs, while the brown components are not. A detailed explanation of each module is provided in Section~\ref{subsec:detailed module description}.

{\bf Inputs.} As inputs, camera images, text prompt, and past ego trajectory are given to the system. Camera images include four most recent sets of images from four cameras (i.e., front, front-zoom, left, and right) at a 10~Hz sampling rate. The text prompt consists of a {\em system prompt} and a {\em user prompt}, where the system prompt defines the identity (or {\em persona}) of the language model. Alpamayo specifically uses the following system prompt:
\halfcolbox{You are a driving assistant that generates safe and accurate actions.}
The following user prompt is also used:
\halfcolbox{Output the chain-of-thought reasoning of the driving process, then output the future trajectory.}
The ego trajectory history is represented as a sequence of 16 poses sampled at 10 Hz over the past 1.6 seconds, expressed in the vehicle-centric coordinate frame defined at the current time step. During inference, the inputs are given in the order of camera images, system prompt, past ego trajectory, and user prompt.

{\bf Outputs.} As outputs of the system, $N$ reasoning sequences and $N$ corresponding future trajectories are generated. By default, Alpamayo generates a single trajectory ($N$=1), which, can be adjusted by modifying a hyperparameter. Each trajectory is converted from an action sequence produced by the action decoder. An action sequence is represented as $\boldsymbol{a} = \{(a^i, k^i)\}_{i=1}^{64}$, where $a^i$ and $k^i$ denote the acceleration and curvature at the $i$-th future timestep, respectively, with each timestep corresponding to $0.1$ seconds (10 Hz). The resulting ego trajectory for the next 6.4~second is expressed as $\boldsymbol{\tau} = \{(x^i, y^i, \theta^i_{\text{yaw}})\}_{i=1}^{64}$, where $(x^i, y^i)$ denotes the two-dimensional position in the ego vehicle's coordinate and $\theta^i_{\text{yaw}}$ denotes the heading angle. Additionally, human readable reasoning messages are generated from the reasoning sequences from the language decoder.


\subsection{Detailed Inference Pipeline}
\label{subsec:detailed module description}

As illustrated in Fig.~\ref{fig:architecture}, the inference pipeline of Alpamayo proceeds through four modules in sequence: preprocessing, reasoning (with vision encoder and language decoder), action generation (with action decoder), and postprocessing. We first describe the single-trajectory case in this section and discuss the multi-trajectory generation in Section~\ref{subsec:generating multiple trajectories}.

\textbf{(i) Preprocessing:} Shown as the yellow rectangle in Fig.~\ref{fig:architecture}, it consists of three components: an image tokenizer, a text tokenizer, and a trajectory tokenizer. During inference, camera images are tokenized by the image tokenizer, which follows the image preprocessor of Qwen3-VL~\cite{Qwen3-VL} and splits each image into patches of $14\times14$ pixels. The text prompt is processed by the text tokenizer, likewise identical to that of Qwen3-VL. Past ego trajectory is handled by the trajectory tokenizer, which extends the text tokenizer with additional vocabularies for discretized trajectory representation, allowing trajectory information to occupy the same token space as textual inputs. The resulting tokens from the image tokenizer are passed to the vision encoder whereas the text and trajectory tokens are concatenated and fed to the text and trajectory encoder.

\textbf{(ii) Reasoning:} Shown as the green rectangle in Fig.~\ref{fig:architecture}, it comprises a vision encoder, a text and trajectory encoder, a language decoder, and a token sampler. The vision encoder, based on Qwen3-VL and composed of a three-dimensional convolutional layer followed by a Vision Transformer with $27$ transformer blocks, maps image tokens into embeddings, which are multi-dimensional vectors. Text and trajectory tokens are simultaneously converted to embeddings by the text and trajectory encoder, implemented as a lookup table. These embeddings are then processed by the language decoder, a decoder-only transformer with $36$ transformer blocks and a linear projection head derived from the Qwen3-VL language model. The token sampler then selects the next token from the probability distribution derived from the logits produced by the language decoder. While the official release~\cite{Alpamayo} refers to the language decoder as the Cosmos-Reason 1~\cite{Cosmos-Reason1} backbone, the two may not share identical weights, as Cosmos-Reason 1 is built on Qwen2.5-VL~\cite{Qwen2.5-VL} whereas the open-source release of Alpamayo is built on Qwen3-VL.

The language decoder operates in two distinct phases: {\em prefill} and {\em decode}. In the prefill phase, all input embeddings are processed in parallel through the $36$ transformer blocks, building an initial understanding of the full input context. In the decode phase, CoT reasoning tokens are generated autoregressively via the language decoder and the token sampler. Each new token is conditioned on all previously generated tokens until a termination token is produced or the maximum generation length is reached. The first token is generated by passing the last hidden state from the prefill phase through the linear projection head and the token sampler. Throughout both phases, each transformer block accumulates KV caches, internal buffers that store intermediate computation results for all previously processed tokens, allowing the model to avoid redundant recomputation when generating subsequent tokens. Upon producing $m$ CoT tokens $\{T_1, T_2, \cdots, T_m\}$, these KV caches serve as the contextual representation that conditions the downstream diffusion-based action generation module. One important observation is that there is no separate interface between the reasoning module and the action generation module. The resulting KV cache content from the reasoning module plays the role of input to the action generation module.

\textbf{(iii) Action generation:} Shown as the red rectangle in Fig.~\ref{fig:architecture}, it consists of two components: an action encoder and an action decoder. The module begins with a randomly  initialized action sequence and refines it over $10$ iterations. In each iteration, the action encoder, a sinusoidal positional encoding followed by a multi-layer perceptron, maps the current action sequence to action embeddings. These embeddings are then processed by the action decoder, which shares the same transformer architecture as the language decoder but with a different hidden dimension, a compatible key and value dimension, and a linear projection head that maps to the action space. Within each transformer block of the action decoder, attention is computed from two distinct sources of keys and values: action-derived key-value pairs computed from the action embeddings themselves, and the KV caches passed from the language decoder. This dual-source attention allows each refinement step to simultaneously integrate action-level context and the full scene understanding encoded during reasoning. Unlike the language decoder, the action decoder processes all action embeddings in a single parallel pass, after which the linear projection head converts each output embedding into an incremental action update. This update, scaled by $0.1$, is added to the current action sequence, and the result becomes the input for the next iteration. After all $10$ iterations, the final action sequence is passed to the postprocessing module.

\textbf{(iv) Postprocessing:} Shown as the gray rounded rectangle in Fig.~\ref{fig:architecture}, it converts generated outputs into interpretable results. It consists of two components: a text detokenizer and an action-to-trajectory converter. The text detokenizer, the inverse of the text tokenizer in the preprocessing module, maps the reasoning token sequence back to human-readable text. The action-to-trajectory converter then transforms the final action sequence into a future ego trajectory based on the unicycle dynamics.

Among the modules, the reasoning and action generation modules contain transformer-based models, whereas the preprocessing and postprocessing modules involve no learned components. For simplicity, positional encodings, which inject token position information into the embeddings, are omitted from Fig.~\ref{fig:architecture}. Specifically, the vision encoder and the language decoder apply positional encodings internally, while the action decoder receives its position encodings externally, computed by the language decoder. DeepStack~\cite{DeepStack} is also omitted, which is a technique that fuses visual features from multiple layers of the vision encoder into the early layers of the language decoder to improve visual understanding during the prefill phase.\par

\label{subsec:Emerging Bottleneck in Single-Reasoning}
\begin{figure}
    \centering
    \begin{subfigure}{\linewidth}
        \centering
        \includegraphics[width=\linewidth]{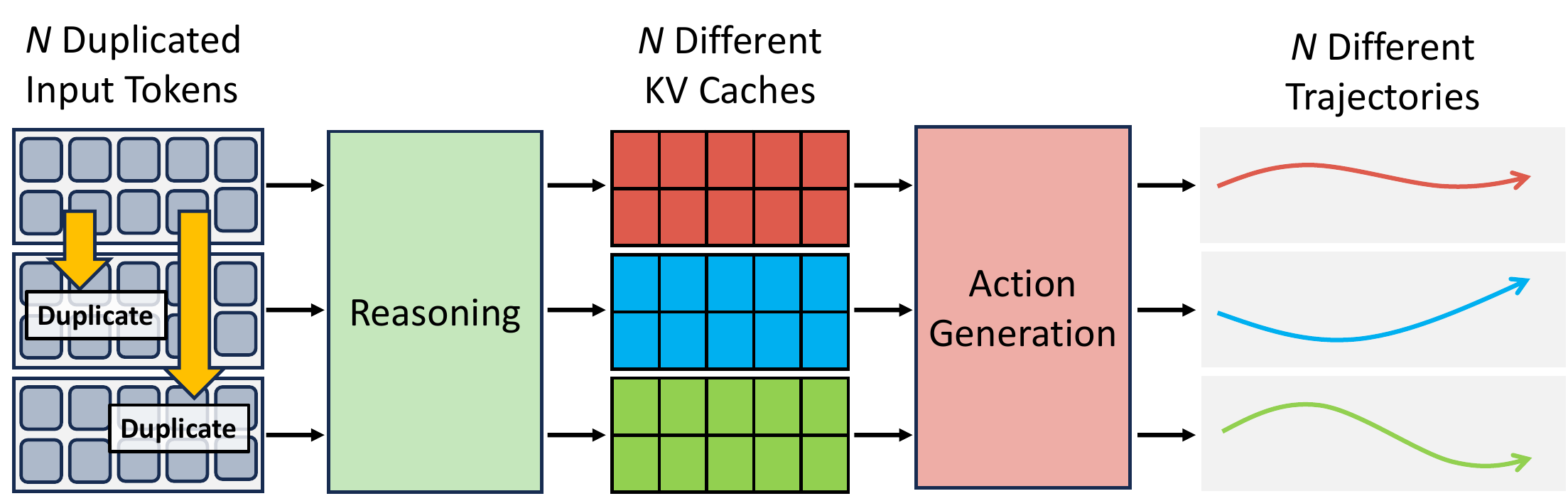}
        \caption{Multi-reasoning architecture}
        \label{fig:multi_reasoning}
    \end{subfigure}

    \begin{subfigure}{\linewidth}
        \centering
        \includegraphics[width=\linewidth]{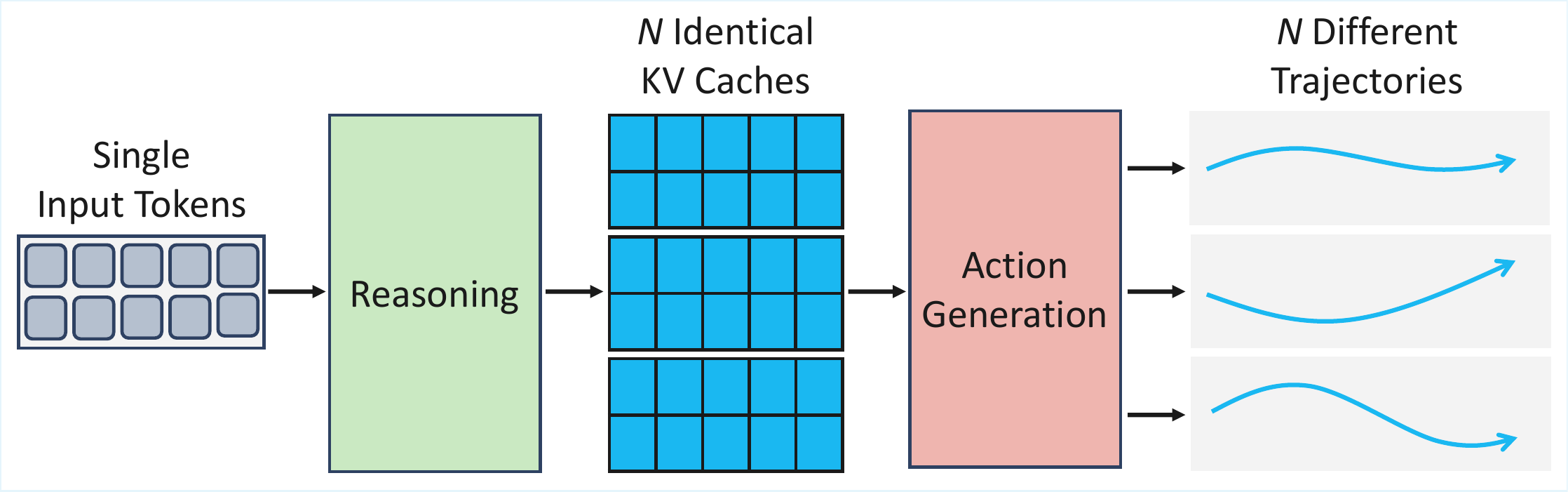}
        \caption{Single-reasoning architecture}
        \label{fig:single_reasoning}
    \end{subfigure}
    
    \caption{Comparison of multi-reasoning and single-reasoning architectures}
    \label{fig:single_multi}
    \vspace{-0.3cm}
\end{figure}

\subsection{Generating Multiple Trajectories}
\label{subsec:generating multiple trajectories}

{\bf Multi-reasoning (N:N).} Alpamayo uses the multi-reasoning approach for generating multiple trajectories. As illustrated in Fig.~\ref{fig:multi_reasoning}, multi-reasoning extends single-trajectory generation to $N$ multi-trajectory generation through batched inference, not iterative inference. The input tokens are replicated to form a batch of size $N$, and the reasoning generation module processes the copies in a batched manner, producing $N$ distinct reasoning sequences and their associated KV caches. The action generation module then operates on a batch of $N$ randomly initialized action sequences, each conditioned on its own KV cache, yielding $N$ trajectories. Trajectory diversity in this approach arises from two sources: stochastic token sampling during reasoning generation, which produces different reasoning sequences and therefore different KV caches, and random initialization of the action sequences in the action generation module.

{\bf Single-reasoning (1:N).} A possible alternative is to use the single-reasoning approach. As illustrated in Fig.~\ref{fig:single_reasoning}, single-reasoning generates a single reasoning sequence and its associated KV cache from the input tokens, then shares this single KV cache across a batch of $N$ randomly initialized action sequences. The action generation module thus produces $N$ trajectories conditioned on the same contextual representation. Trajectory diversity in this approach relies solely on the random initialization of action sequences, since the reasoning sequence and its KV cache are identical for all $N$ trajectories. The diversity advantage of multi-reasoning over single-reasoning therefore hinges on how strongly variation in the reasoning sequence influences the downstream action generation. If this influence is limited, the stochasticity of action sequence initialization alone may be sufficient to achieve comparable trajectory diversity.

\section{Architecture Redesign by Latency Analysis}
\label{sec:delay component}

\subsection{Profiling Setup}
\label{setup}

For the latency component profiling of Alpamayo, we use an NVIDIA DGX Spark machine with a GB10 Grace Blackwell processor and 128~GB of integrated memory. As the operating system, we use NVIDIA DGX OS 7, which is based on Ubuntu~24.04 with preinstalled GPU drivers and CUDA runtime. As an evaluation dataset, a random scenario with 16 cameras images is selected from NVIDIA Physical AI Dataset. Each measurement result represents the average value of 100 measurement iterations. For profiling, we modify Alpamayo inference code and Hugging Face Transformers library by injecting profiling code to precisely capture each module's latency.

By the profiling results, Alpamayo's inference latency can be analyzed by five major latency components, denoted as follows:
\begin{itemize}
\item{\bf Preprocessing:} Latency caused by the preprocessing module, which is mostly for input token generations from inputs.
\item{\bf Reasoning-Vision:} Latency caused by the vision encoder in the reasoning module, which is mostly about executing ViT-based transformer blocks for image encoding.
\item{\bf Reasoning-Prefill:} Latency caused by the prefill phase by the language decoder in the reasoning module, which populates KV caches for subsequent token generations.
\item{\bf Reasoning-Decode:} Latency caused by the autoregressive CoT tokens generations during the decode phase in the reasoning module; this latency grows with the number of generated tokens.
\item{\bf Action-Gen:} Latency caused by the iterative refinement of action sequences by the action decoder in the action generation module, which runs for 10 iterations.
\end{itemize} 
Since the postprocessing module's latency is negligible without visible impact on the inference latency, we do not separately report the postprocessing latency. We report the profiling results for the multi-reasoning and single-reasoning cases in Section~\ref{sec:mr} and Section~\ref{sec:sr}, respectively.
Then Section~\ref{sec:num_token} demonstrates the relation between the action generation latency and the number of generated CoT tokens.

\begin{figure}[t]
    \centering
    \begin{subfigure}[t]{0.48\columnwidth}
        \centering
        \includegraphics[width=\linewidth]{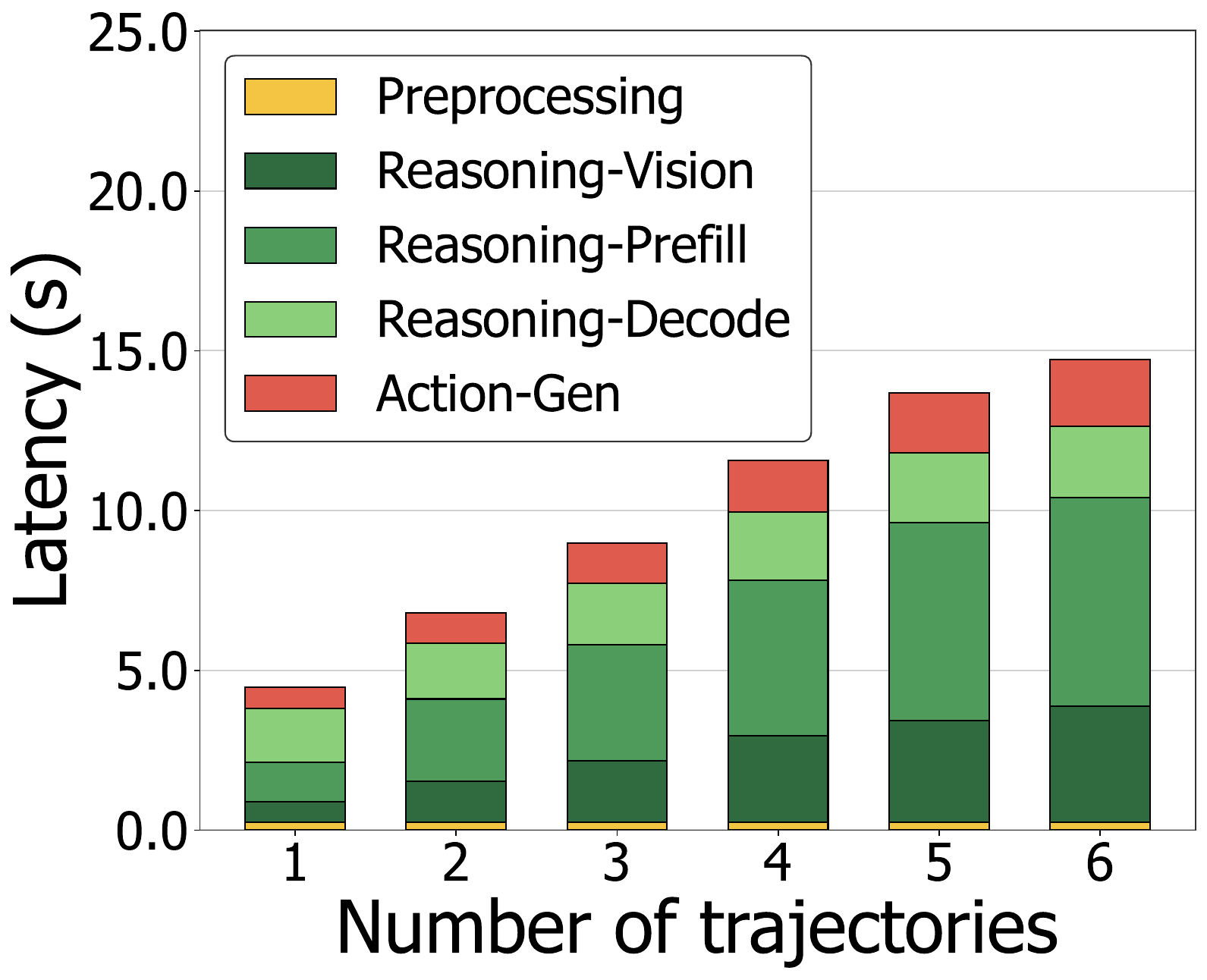}
        \caption{Latency breakdown}
        \label{fig:multi_breakdown}
    \end{subfigure}
    \hfill
    \begin{subfigure}[t]{0.48\columnwidth}
        \centering
        \includegraphics[width=\linewidth]{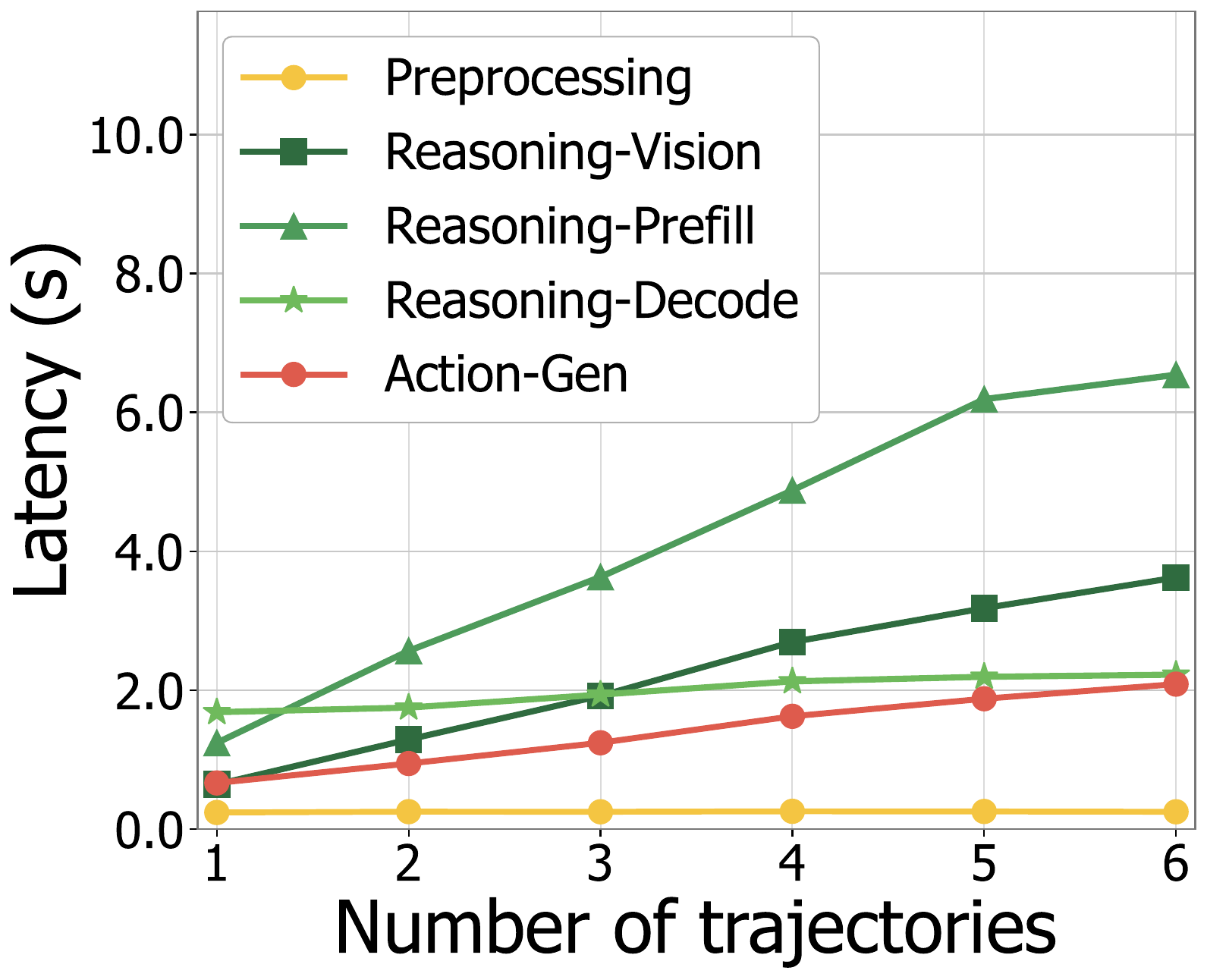}
        \caption{Latency trends}
        \label{fig:Multi_latency}
    \end{subfigure}
    
    \caption{Latency analysis for multi-reasoning architecture}
    \label{fig:multi}
        \vspace{-0.3cm}
\end{figure}

\label{sec:e2e_latency_single_reasoning}
\begin{figure}
    \centering
    \begin{subfigure}[t]{0.48\columnwidth}
        \centering
        \includegraphics[width=\linewidth]{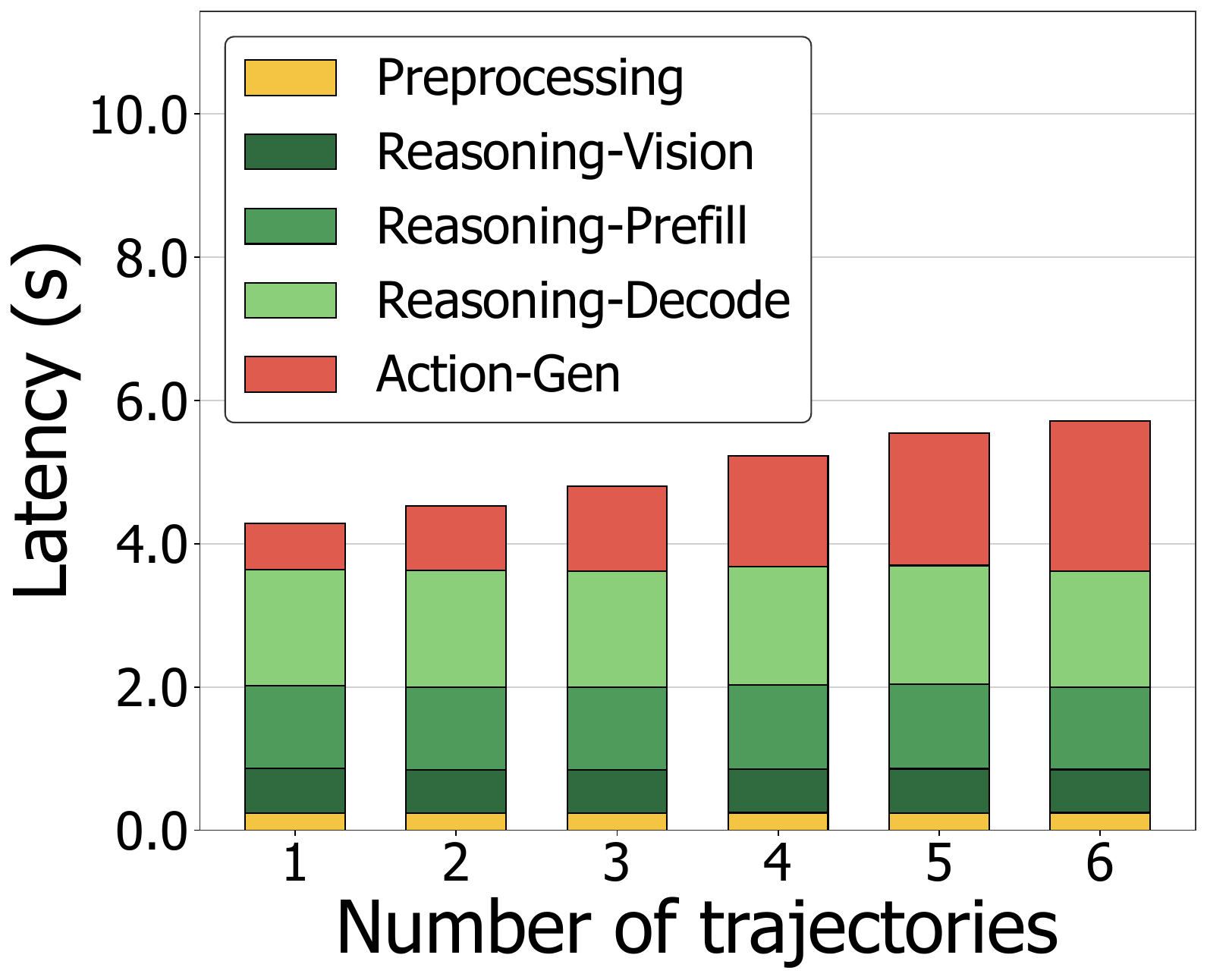}
        \caption{Latency breakdown}
        \label{fig:single_breakdown}
    \end{subfigure}
    \hfill
    \begin{subfigure}[t]{0.48\columnwidth}
        \centering
        \includegraphics[width=\linewidth]{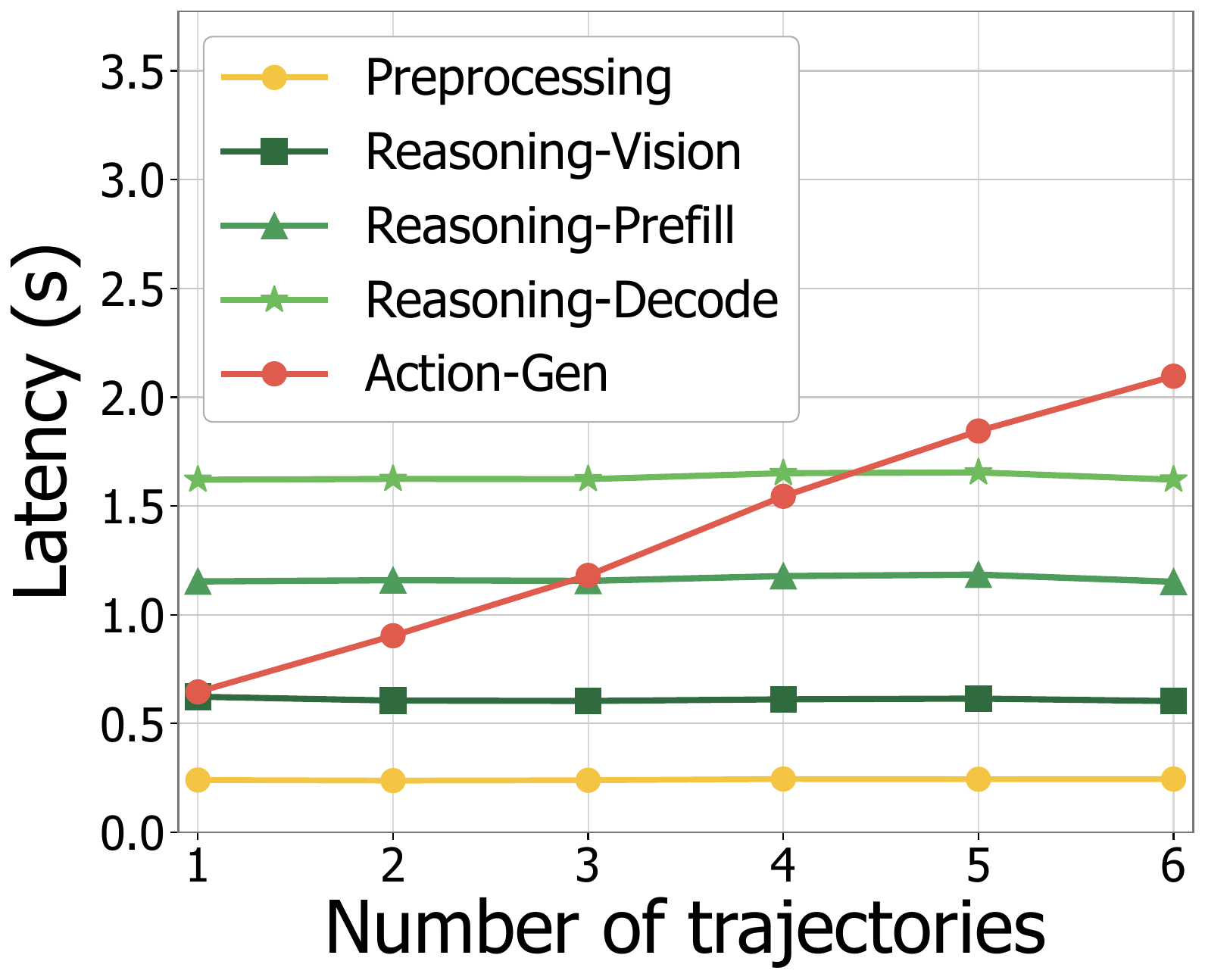}
        \caption{Latency trends}
        \label{fig:single_latency}
    \end{subfigure}
    \caption{Latency analysis for single-reasoning architecture}
    \label{fig:single}
    \vspace{-0.3cm}
\end{figure}

\subsection{Alpamayo's Multi-Reasoning Architecture}
\label{sec:mr}

Fig.~\ref{fig:multi_breakdown} shows the per-component latency as a stacked bar graph for varying numbers of trajectories from 1 to 6, using Alpamayo's default {\em multi-reasoning} architecture. Fig.~\ref{fig:Multi_latency} shows the same results broken down by individual latency component to highlight each component's trend.

The preprocessing latency remains constant at around 0.25~s regardless of the number of trajectories, indicating that input duplication occurs after the preprocessing module. In contrast, the reasoning latency increases almost linearly with the number of trajectories, as the duplicated inputs are processed in a batched manner, scaling the computational workload proportionally. However, by looking at Fig.~\ref{fig:Multi_latency}, we can notice that the slopes are different for each latency component. More specifically, the scaling factor of Reasoning-Vision is 5.63, and that of Reasoning-Prefill is 5.28. However, the scaling factor of Reasoning-Decode is just 1.32. This discrepancy indicates the different computational intensity of each component, meaning Reasoning-Decode requires far less computations than Reasoning-Vision and Reasoning-Prefill. The Action-Gen latency also scales by 3.15, meaning that its computation density is in between Reasoning-Vision and Reasoning-Decode.


\begin{figure}[t]
    \centering
    \includegraphics[width=\linewidth]{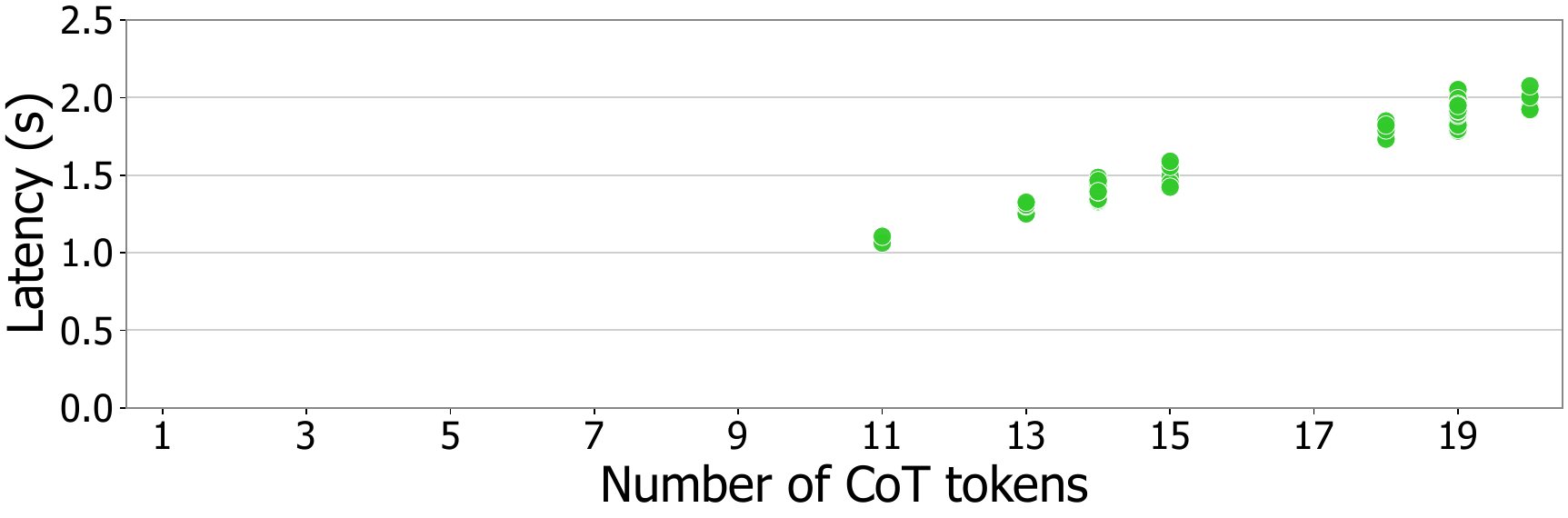}
    \caption{Relation between number of CoT tokens and decoding latency}
    \label{fig:decode}
    \vspace{-0.3cm}
\end{figure}

\begin{figure*}[!t]
    \centering
    \begin{subfigure}{\linewidth}
        \centering
        \includegraphics[width=\linewidth]{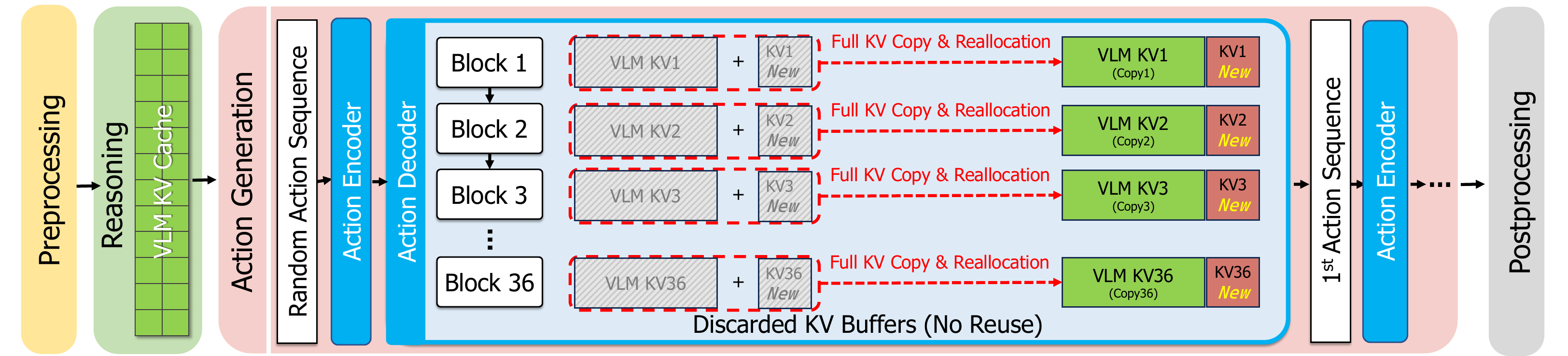}
        \caption{Baseline action generation architecture with {\em dynamic} KV cache management scheme}
        \label{fig:dynamic}
    \end{subfigure}
    
    \vspace{0.5em}
    
    \begin{subfigure}{\linewidth}
        \centering
        \includegraphics[width=\linewidth]{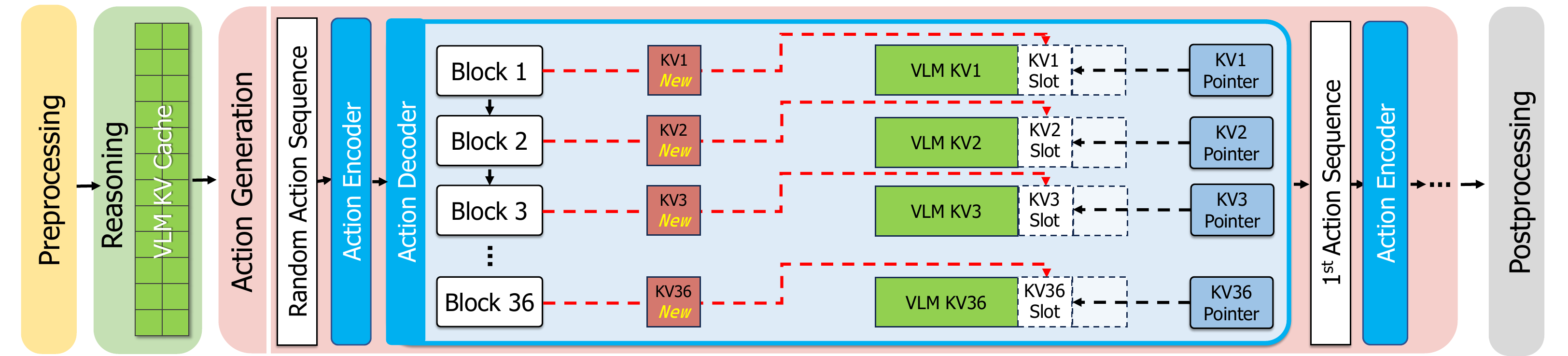}
        \caption{Our optimized action generation architecture with {\em static} KV cache management scheme}
        \label{fig:static}
    \end{subfigure}
    
    \caption{Optimizing action generation}
    \label{fig:action-gen}
\end{figure*}

{\bf Discussion.} Note that even in a very powerful DGX Spark machine, the inference latency for the single trajectory generation is about 4~s, which is not practical for real-time processing. However, this does not mean that Alpamayo is not useful in real-world autonomous driving. VLA-based models are often concurrently used in combination with lightweight E2E models for safety regulation purposes~\cite{DriveVLM, Senna}, where high inference latencies are acceptable. Even in such cases, however, the linearly scaling latency with the number of trajectories remains a critical concern.

\subsection{Our Single-Reasoning Architecture}
\label{sec:sr}


To overcome the limitations of the multi-reasoning architecture, we modify Alpamayo to adopt a {\em single-reasoning} architecture, eliminating input duplication so that the reasoning module generates a single reasoning sequence shared across all trajectories. However, to generate multiple trajectories, identical KV caches from the single reasoning are used to generate multiple trajectories. Thus, the action generation module runs batched inference as of the multi-reasoning architecture. Even with the same KV cache, diverse trajectories can still be generated owing to the random initialization of the action sequence at the beginning of the action generation process.

Fig.~\ref{fig:single} shows the profiling results in our single-reasoning implementation, where it begins with the same latency as the multi-reasoning case when the number of trajectories is 1. However, as the number of trajectories increases, all the latency components other than the action generation latency are constant, meaning that there is no additional computation overhead affected by the number of trajectories. Only the action generation latency scales by 3.15, which is the same as the multi-reasoning case. By looking at Fig.~\ref{fig:single_breakdown}, the proportion of action generation is only 15.06~\% when the number of trajectories is 1. However, When the number of trajectories is 6, the proportion reaches 33.67~\%, which is significant within the overall latency.


{\bf Discussion.} It is difficult to do an apples-to-apples comparison between the multi-reasoning architecture and the single-reasoning architecture. However, it is obvious that latency overhead caused by the multi-reasoning architecture is critical when we need a large number of trajectories. Thus, on the condition that the trajectory diversity and the resulting driving performance remains with little changes, the single-reasoning architecture can be a better solution.

\subsection{Decoding Latency}
\label{sec:num_token}

Fig.~\ref{fig:decode} shows the decode latency for varying numbers of generated CoT tokens. The figure shows that decode latency increases linearly with the number of generated tokens since the tokens per second (TPS) performance is invariant with a given language model on a fixed hardware platform, which is 10.12 TPS in this case. The maximum token generation length of Alpamayo is 256 by default, but in practice it generates between 11 and 20 CoT tokens, limiting the latency variation to a narrow range. To guarantee a hard decode latency, we may impose a less number of output tokens than the default one. Nonetheless, limiting output tokens may meaningfully affect the reasoning capability of language models~\cite{s1STTS, ScalingLLM}. We therefore maintain the default limit, but we have not experienced more than 20 CoT tokens yet. This output token limit issue is outside the scope of this study. 

\section{Optimizing Action Generation}
\label{sec:optimizing}


\subsection{Baseline Action Generation Architecture}
\label{subsec:depth_analysis}

Fig.~\ref{fig:dynamic} shows Alpamayo's baseline action generation architecture. It begins with a random action sequence, which is refined with 10 diffusion iterations. Each diffusion iteration is composed of 36 transformer blocks. Action transformer blocks reference the KV cache generated by their corresponding transformer blocks in the reasoning module to incorporate reasoning context into action generation. At the same time, they produce new KV entries that are appended to the corresponding KV caches. This block-wise KV cache referencing and updating process is repeated across all transformer blocks. After Block 36, a new diffusion iteration begins using the reasoning module’s KV cache together with the action sequence output from the previous iteration. This execution pattern continues until the final diffusion iteration.


Alpamayo adopts a {\em dynamic} KV cache management scheme, where each transformer block copies the previously accumulated KV cache (full KV) concatenated with newly generated KV tensors to a newly allocated buffer. This dynamic scheme is commonly used in conventional LLM applications like chatbots, as the output sequence length is non-deterministic, and thus the maximum KV cache size cannot be determined in advance. While dynamically expanding the KV cache is a scalable solution in such cases, it introduces significant memory copy and reallocation overhead.




\subsection{Optimization 1: Static KV Cache Management} 
\label{subsec:Exploiting Opportunity 1: Copy And Allocation}

\begin{figure*}[t]
    \centering
    \includegraphics[width=\linewidth]{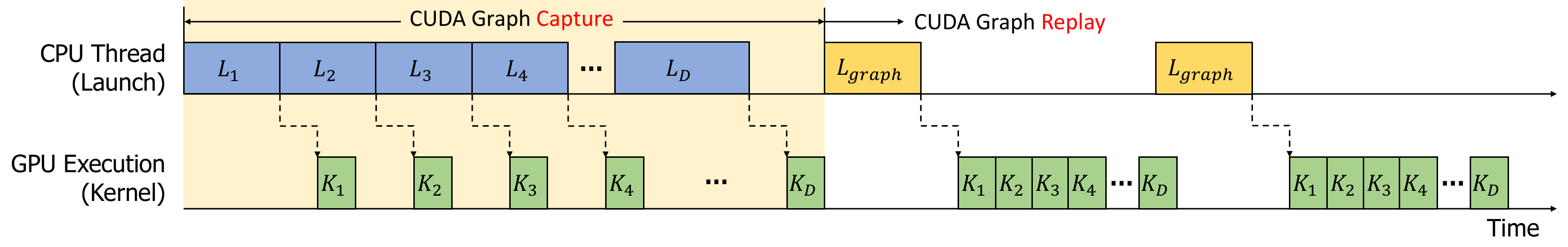}
    \caption{CUDA graph capture}
    \label{fig:graph}
\end{figure*}

To address the inefficiency of dynamic KV cache management, we leverage a key characteristic of the action generation module: its memory usage becomes predictable after the reasoning stage. Although the KV cache produced by the VLM cannot be determined in advance, it remains unchanged during diffusion. Furthermore, the KV entries generated during diffusion are identical in size across iterations. Specifically, the KV cache received from the language decoder occupies approximately 454.3~MB for a single trajectory when assuming 20 CoT tokens. During action generation, the KV cache increases by 9.44~MB, reaching a maximum size of 463.74~MB of KV cache. Based on this observation, we implement a {\em static} KV cache management scheme that preallocates the maximum required memory in advance and updates the KV cache via in-place writes. Specifically, we maintain a KV pointer that tracks the correct memory address for each transformer block and advances through the preallocated buffer at each diffusion iteration, as illustrated by the blue box in Fig.~\ref{fig:static}. This approach eliminates unnecessary copy-and-allocation operations in the baseline dynamic KV cache management scheme.




\subsection{Optimization 2: Removing GPU Kernel Launch Overhead}
\label{subsec:CUDA graph capture}

Through our static KV cache management, both the memory address and size of the KV cache are fixed across diffusion iterations. Since the same GPU kernels are executed at every iteration, each kernel references identical memory addresses across iterations, as every iteration is identical in terms of input and tensor dimensions.

This deterministic pattern of GPU kernel execution and memory access enables an additional optimization: recording and replaying GPU kernel launches for diffusion to reduce diffusion latency, leveraging CUDA graphs as illustrated in Fig.~\ref{fig:graph}. As shown on the left-hand side, a CPU thread launches $D$ GPU kernels sequentially. Each launch incurs kernel launch overhead (denoted as $L_1, L_2, \dots, L_D$), corresponding to the time required to issue kernel execution requests to the GPU. The dotted lines represent the delay between kernel launch and actual GPU execution, which often results in GPU idle periods. These gaps become more pronounced when the kernel execution time is shorter than the launch overhead, leading to significant underutilization of GPU resources. However, as shown on the right-hand side of Fig.~\ref{fig:graph}, we capture the entire launch sequence and replay it as a single $L_{\text{graph}}$ using CUDA graphs. This eliminates per-kernel launch overhead and the associated idle gaps, thereby improving overall GPU utilization.

\begin{figure*}[t]
    \centering

    \begin{subfigure}{\textwidth}
        \centering
        \begin{subfigure}{0.24\textwidth}
            \includegraphics[width=\linewidth]{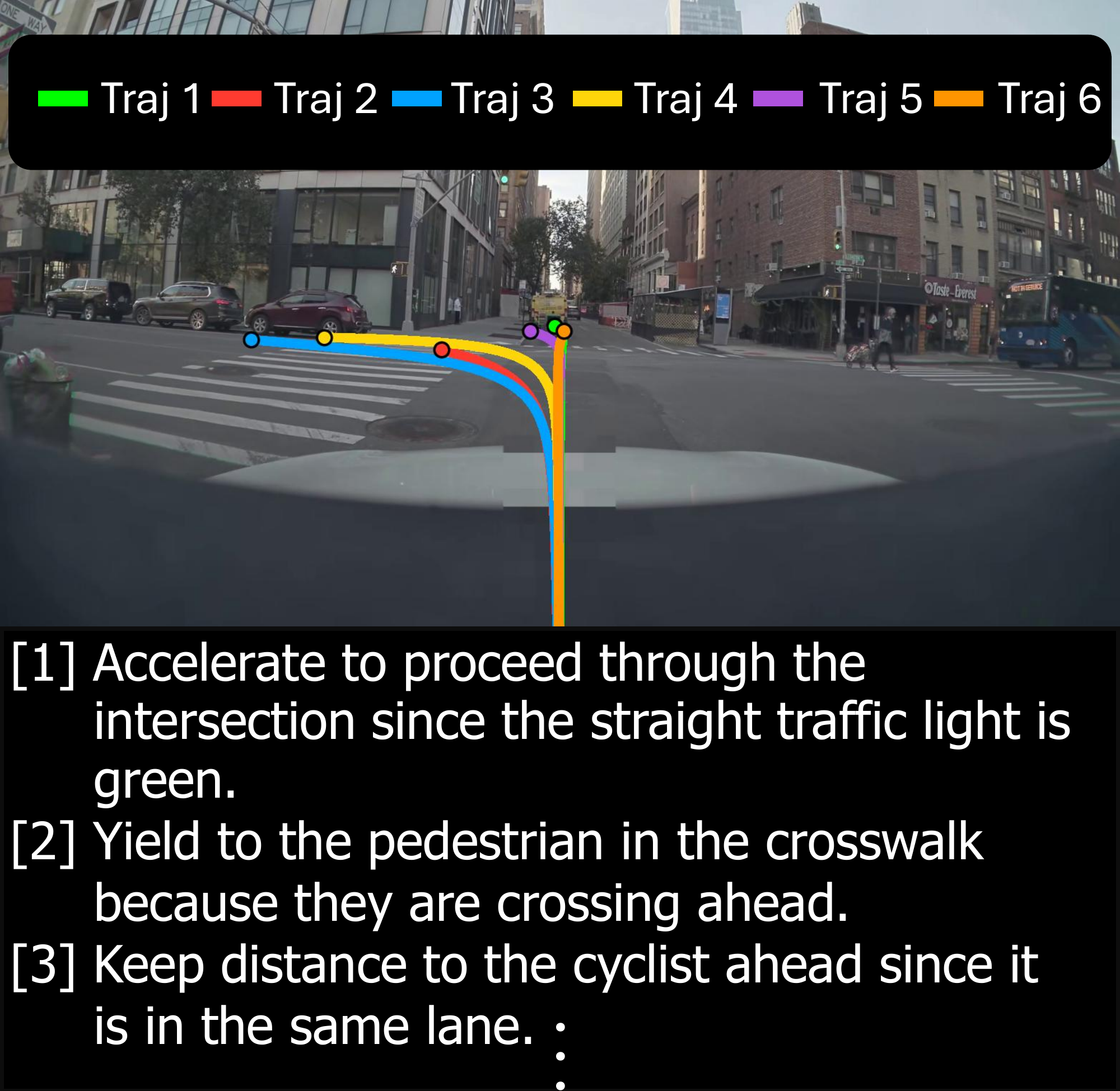}
        \end{subfigure}
        \hfill
        \begin{subfigure}{0.24\textwidth}
            \includegraphics[width=\linewidth]{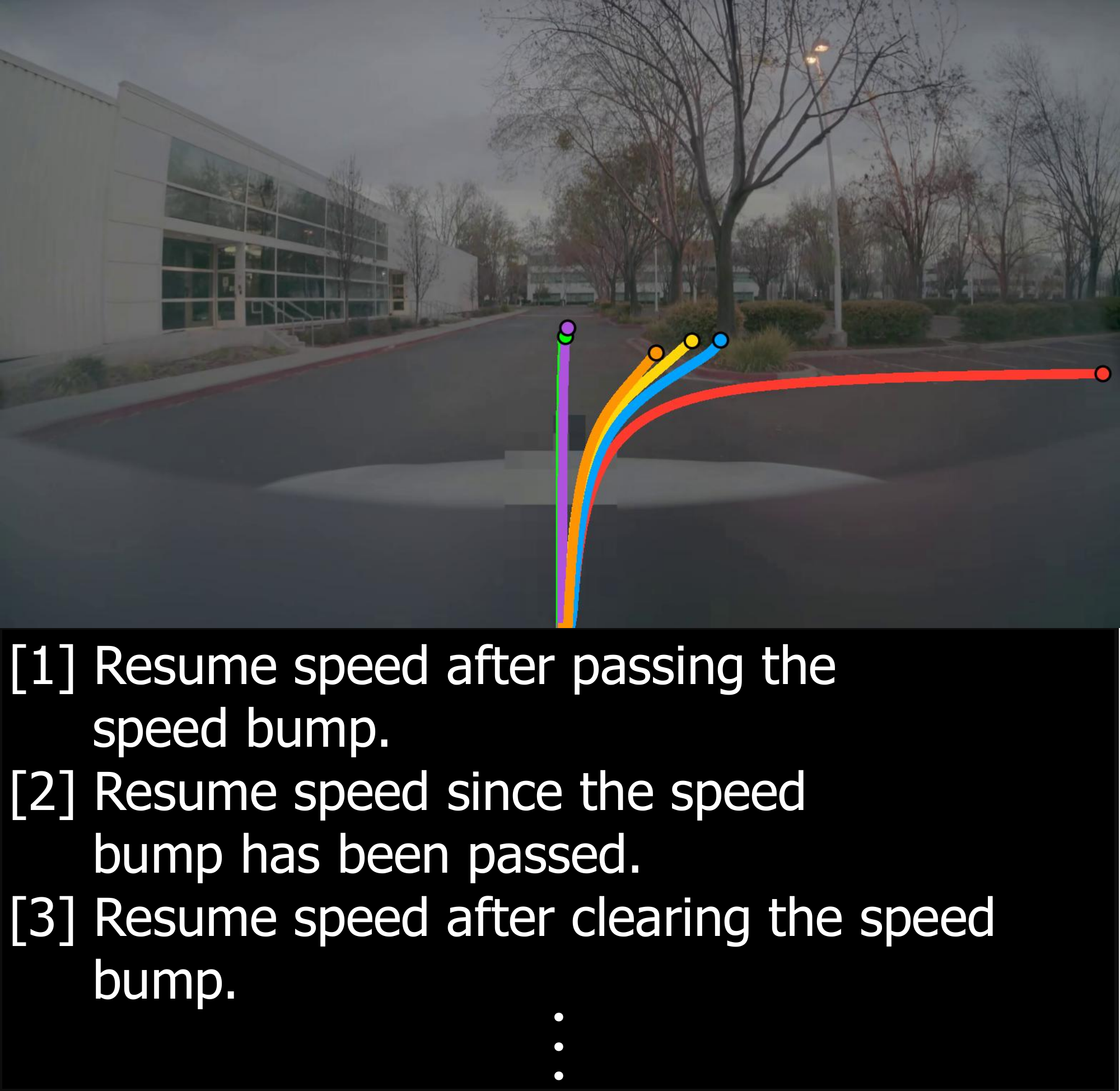}
        \end{subfigure}
        \hfill
        \begin{subfigure}{0.24\textwidth}
            \includegraphics[width=\linewidth]{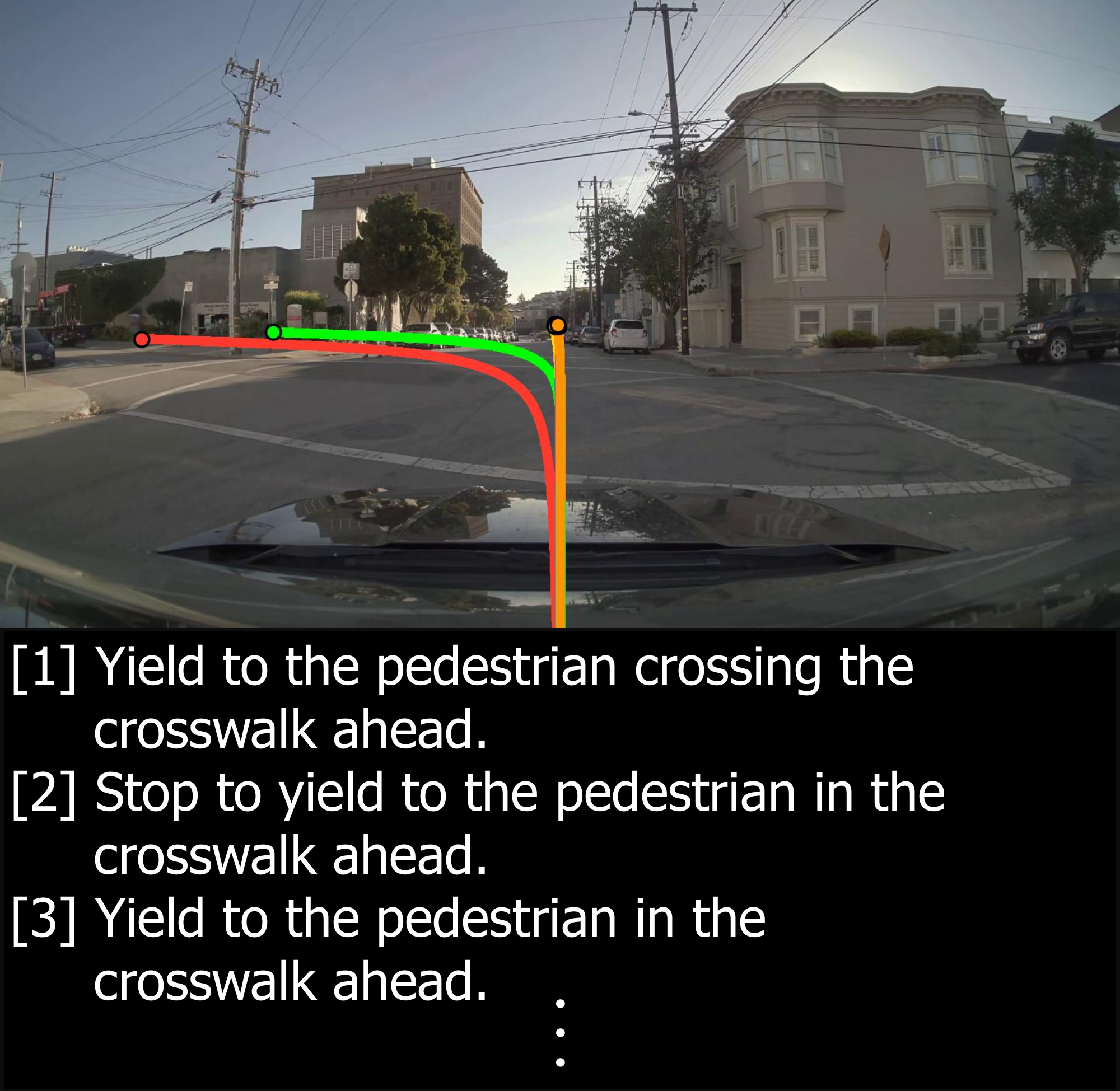}
        \end{subfigure}
        \hfill
        \begin{subfigure}{0.24\textwidth}
            \includegraphics[width=\linewidth]{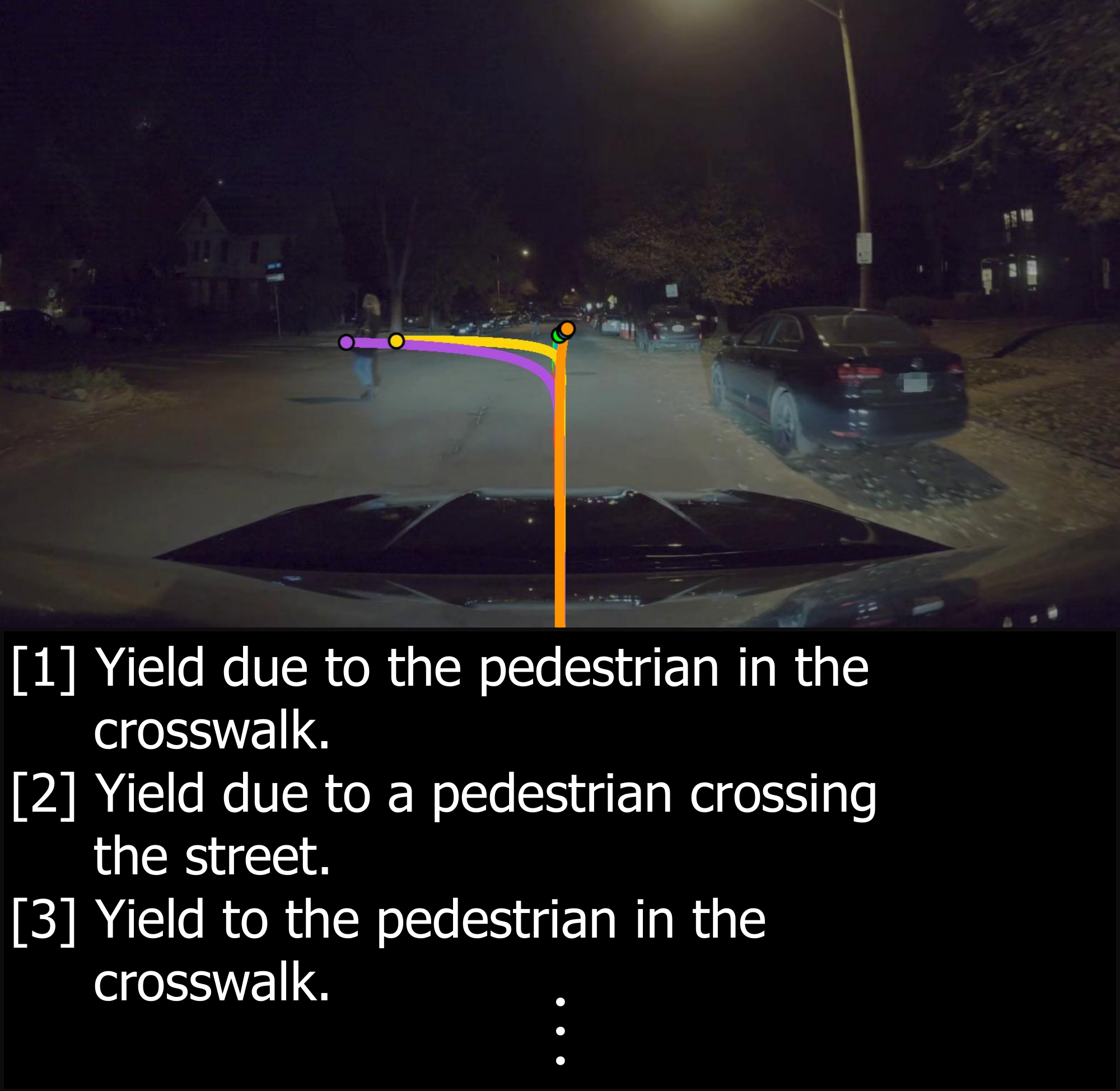}
        \end{subfigure}
        
        \caption{Multi-reasoning results}
    \end{subfigure}

    \vspace{0.6em}

    \begin{subfigure}{\textwidth}
        \centering
        \begin{subfigure}{0.24\textwidth}
            \includegraphics[width=\linewidth]{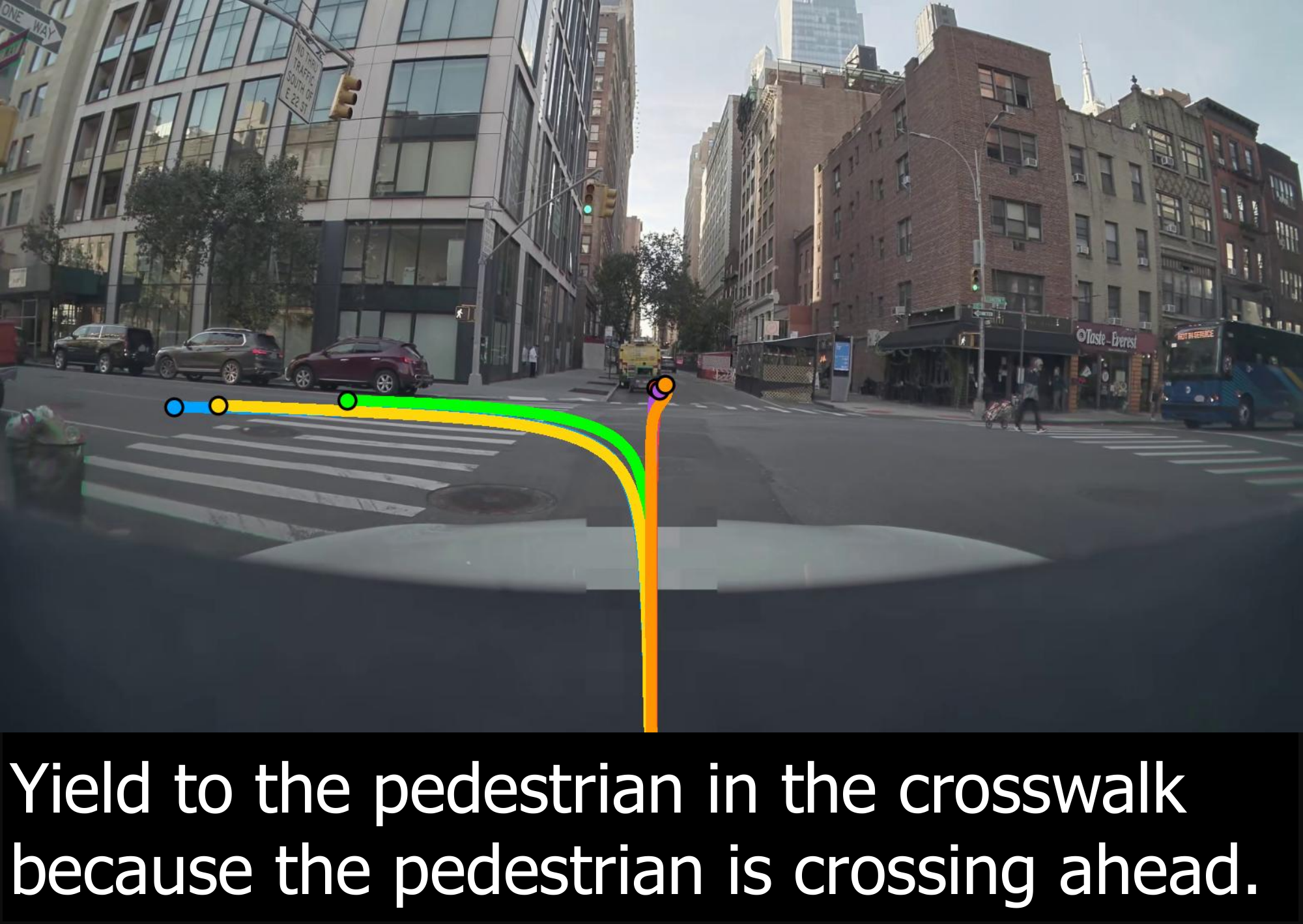}
        \end{subfigure}
        \hfill
        \begin{subfigure}{0.24\textwidth}
            \includegraphics[width=\linewidth]{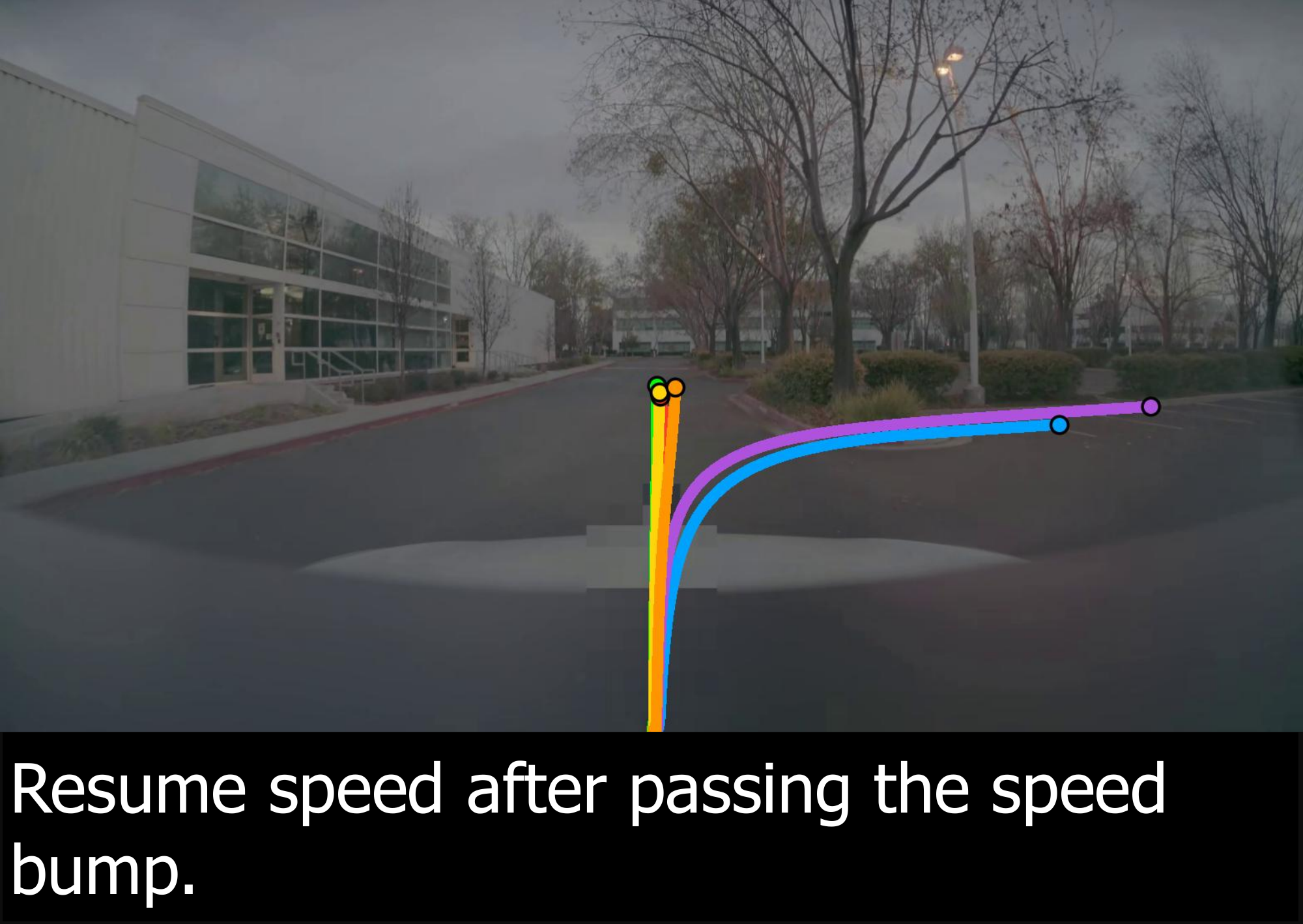}
        \end{subfigure}
        \hfill
       \begin{subfigure}{0.24\textwidth}
            \includegraphics[width=\linewidth]{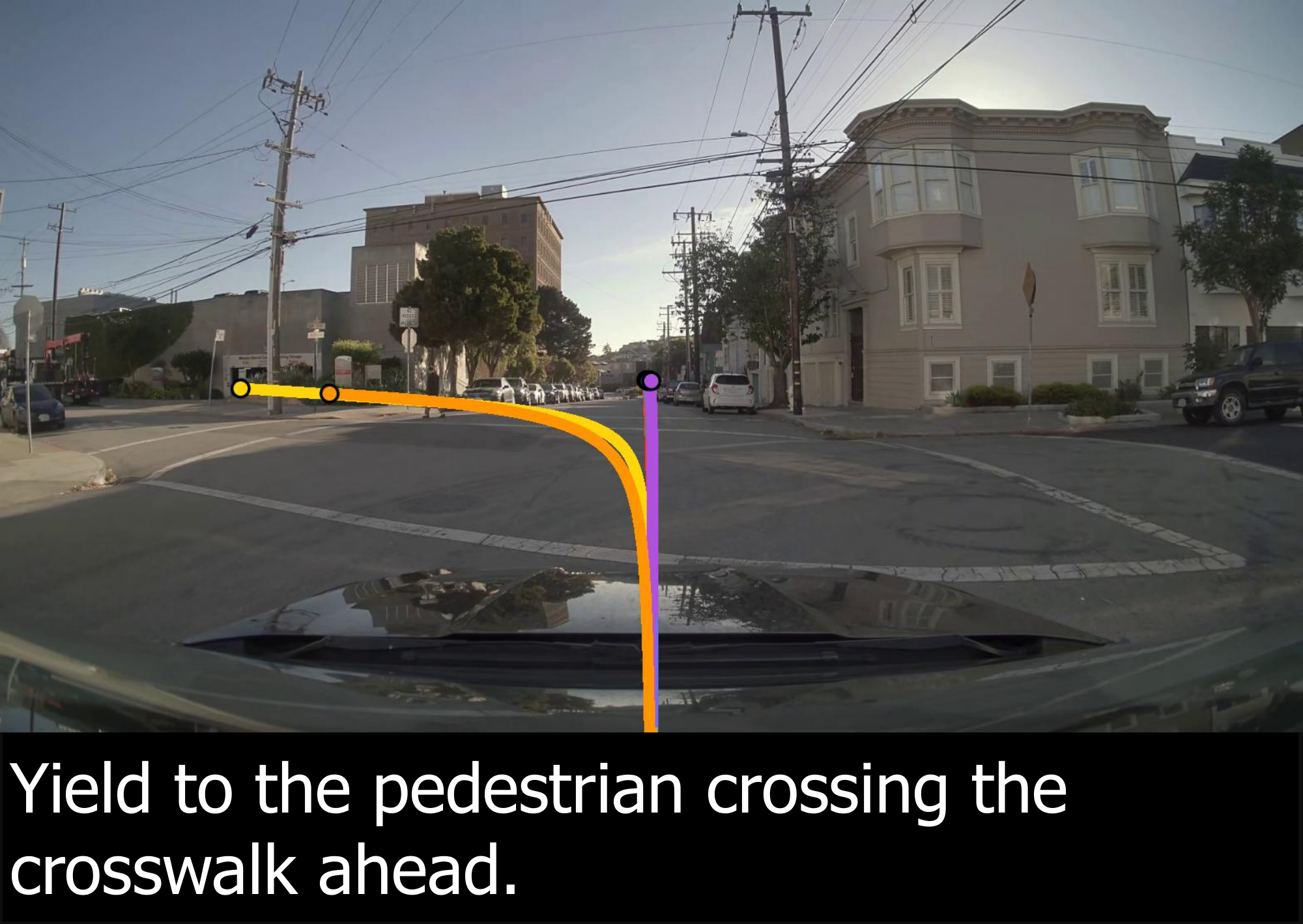}
        \end{subfigure}
        \hfill
        \begin{subfigure}{0.24\textwidth}
            \includegraphics[width=\linewidth]{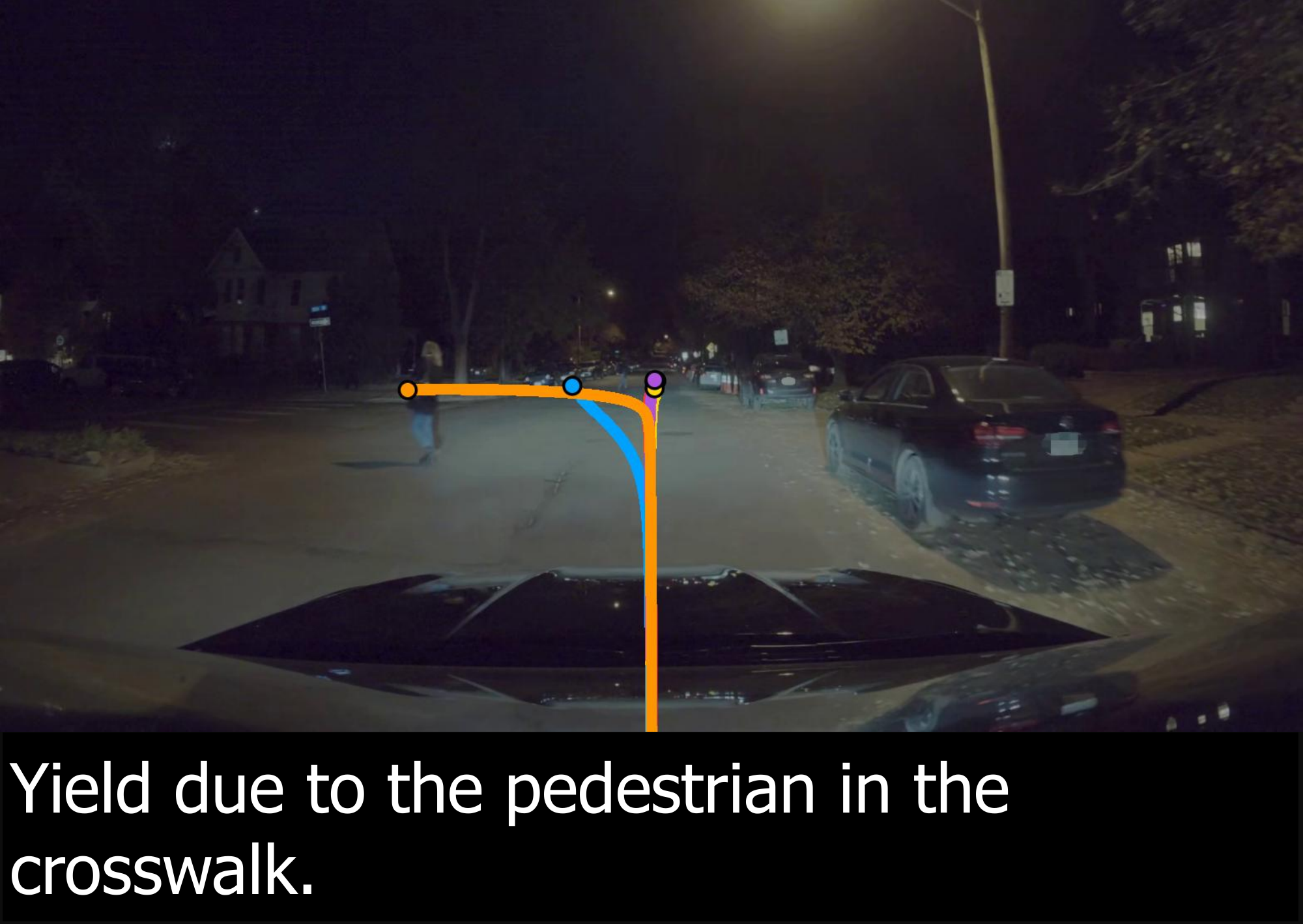}
        \end{subfigure}
        
        \caption{Single-reasoning results}
    \end{subfigure}

    \caption{Trajectory diversity comparison between multi-reasoning and single-reasoning with six trajectories with CoT messages}
    \label{fig:diversity}
\end{figure*}

\section{Experiments}
\label{sec:experiments}

\subsection{Implementation}

{\bf Baseline.} Our implementation builds upon the Alpamayo inference codebase. We do not modify model parameters, as our contributions are confined to the inference architecture, specifically targeting batch processing and KV cache management. We modify the Alpamayo inference code to apply our optimizations, while the Hugging Face Transformers library is modified solely for profiling purposes. We also note that Alpamayo 1.5~\cite{alpamayo1.5_github} was recently released, extending Alpamayo with navigation-conditioned reasoning, post-reinforcement learning, and other features unrelated to inference performance. In our preliminary evaluation, Alpamayo 1.5 demonstrates improved trajectory prediction compared to the original Alpamayo. However, since both models share the same inference architecture, our optimization techniques can be directly applied to Alpamayo 1.5 without modification.

{\bf Single-reasoning architecture.} The official Alpamayo inference code provides a hyperparameter named \texttt{num\_traj\_samples}, to control the number of generated trajectories. However, this approach replicates the input to the reasoning module along the batch dimension for each trajectory, resulting in redundant reasoning generation. We avoid this by setting \texttt{num\_traj\_samples} to 1. Once the reasoning generation completes, the resulting KV cache is duplicated along the batch dimension to match the desired number of trajectories. The expanded KV cache is then passed to the trajectory generation module to generate multiple trajectories. This modification enables efficient single-reasoning, multi-trajectory generation while preserving the model parameters.


{\bf Optimized action generation.} We apply two optimizations to the action generation module, targeting KV cache management and GPU kernel execution. First, we replace dynamic KV cache management with our static scheme. The original \texttt{torch.cat}-based KV cache updates are eliminated; instead, memory is preallocated prior to the diffusion process, and the cache is updated via index-based in-place assignment. This enables the reuse of memory space for KV cache without unnecessary memory allocations. Second, we incorporate CUDA graphs to capture and replay kernel launches across diffusion iterations. Graph capture is performed during the second iteration, after which subsequent iterations are executed via CUDA graph replay, significantly reducing the kernel launch overhead.

\begin{table}
    \centering
    \caption{Trajectory prediction performance comparison between multi-reasoning and single-reasoning}
    \label{table:quality}
    
    \begin{tabular}{lcc}
        \toprule
        \multirow{2}{*}{\textbf{Architecture}} 
        & \multicolumn{1}{c}{\textbf{Open-Loop}} 
        & \multicolumn{1}{c}{\textbf{Closed-Loop (AlpaSim)}} \\
        \cmidrule(lr){2-2} \cmidrule(lr){3-3}
        & \textbf{minADE$_6$@6.4s} 
        & \textbf{DTF (m)} \\
        \midrule
        Multi-reasoning & 0.725 & 131.58 \\
        Single-reasoning & 0.743 & 137.12 \\
        \bottomrule
    \end{tabular}
    \label{tab:prediction}

\end{table}


\subsection{Evaluation Results}

{\bf Trajectory diversity.} The latency analysis comparing the single-reasoning and multi-reasoning architectures has already been presented in Section~\ref{sec:delay component}. In addition, Fig.~\ref{fig:diversity} provides a visual comparison of trajectory diversity using four representative scenarios from the NVIDIA Physical AI Dataset. Although the two approaches do not produce identical trajectories, the single-reasoning results are visually similar to those of multi-reasoning in terms of trajectory diversity. Similar trends are consistently observed across other scenarios beyond the four shown in Fig.~\ref{fig:diversity}. This similarity indicates that randomly initialized action sequences provide sufficient diversity for multi-trajectory generation, whereas generating multiple independent reasonings often produces overlapping results and fails to provide adequate diversity.


{\bf Trajectory prediction.} Since diversity alone does not guarantee the autonomous driving performance, we compare the open-loop and closed-loop trajectory prediction benchmark results in Table~\ref{tab:prediction}. First, we conduct an open-loop evaluation to measure prediction accuracy. The evaluation follows the same setup as \cite{Alpamayo} and uses minADE6@6.4s as the evaluation metric. The dataset consists of 394 scenarios, each with a duration of 20 seconds. For each scenario, we generate six trajectory samples and compute the minimum average displacement error (ADE) among them over a 6.4-second prediction horizon. The results show that the proposed single-reasoning approach achieves a minADE6@6.4s score of 0.743, which is comparable to 0.725 obtained by the multi-reasoning approach, indicating no significant difference in prediction accuracy. Next, we evaluate driving stability in a closed-loop setting using the AlpaSim simulator. We adopt the Distance-to-Failure (DTF) as the evaluation metric, which we define by ourselves due to the lack of explicit details regarding the closed-loop evaluation metric used in \cite{Alpamayo}. Specifically, DTF measures the distance traveled by the ego vehicle before the first failure event occurs. A failure event is defined as one of the following: (i) a collision with surrounding traffic participants, (ii) leaving the drivable area, or (iii) a lateral deviation exceeding 4~m from the ground-truth trajectory. This metric reflects the robustness of generated trajectories in interactive driving scenarios. The closed-loop results show that the proposed single-reasoning approach achieves 137.12~m, slightly better than 131.58~m obtained by the multi-reasoning approach, indicating no performance degradation in driving stability.

\begin{figure}
    \centering
    \includegraphics[width=\columnwidth]{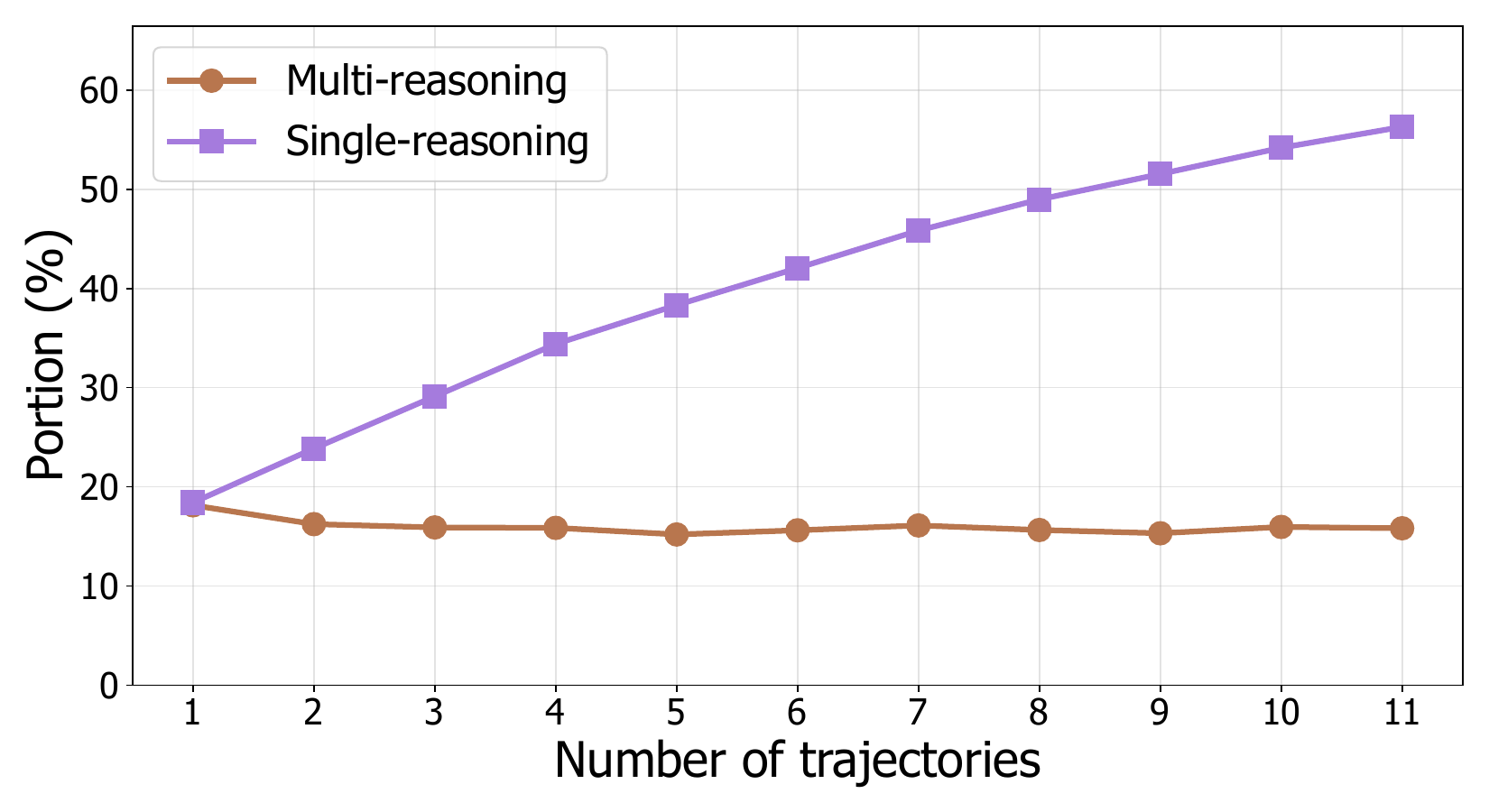}
    \caption{Proportion of action generation latency with varying numbers of trajectories}
    \label{fig:combined}
\end{figure}

\begin{figure}
    \centering
    \begin{subfigure}[t]{0.48\columnwidth}
        \centering
        \includegraphics[width=\linewidth]{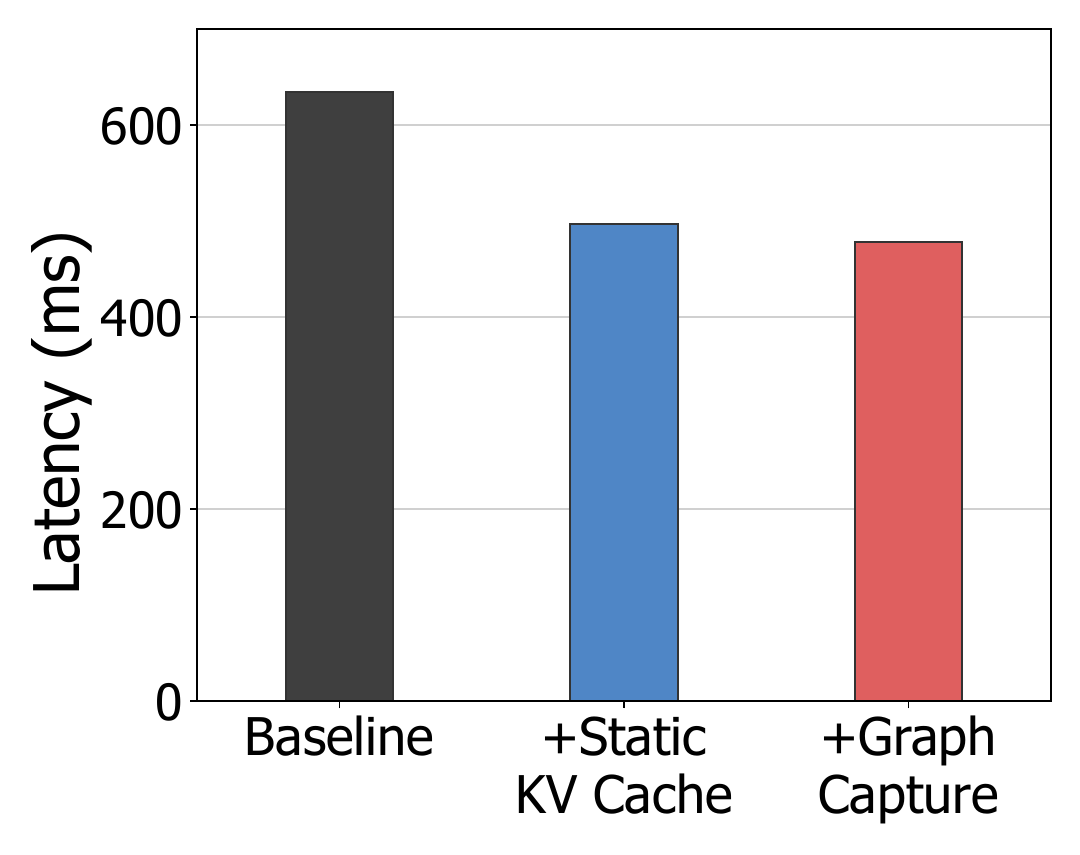}
        \caption{Latency comparison by applying optimization techniques}
        \label{fig:diff_overview}
    \end{subfigure}
    \hfill
    \begin{subfigure}[t]{0.48\columnwidth}
        \centering
        \includegraphics[width=\linewidth]{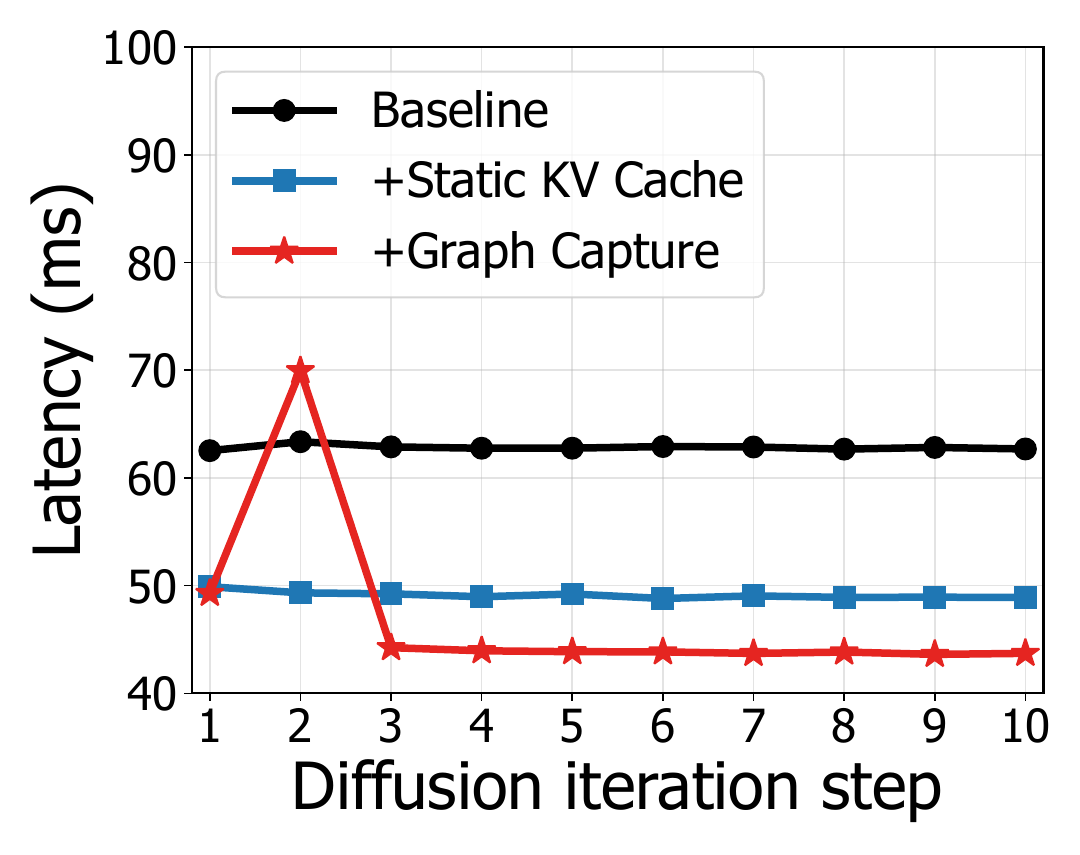}
        \caption{Per-iteration comparion}
        \label{fig:diff_step}
    \end{subfigure}
    
    \caption{Action generation latency optimization results}

    \label{fig:diff_combined}
\end{figure}

{\bf Why efficient action generation is important.} Fig.~\ref{fig:combined} shows the proportion of action generation latency within the overall inference latency as the number of trajectories increase, both multi- and single-reasoning architectures. When generating a single trajectory, the contribution of action generation is nearly identical between the two architectures, accounting for 18.13~\% in multi-reasoning and 18.41~\% in single-reasoning, respectively. However, as the number of trajectories increases, the contribution grows significantly in the single-reasoning architecture, while it remains stable in the multi-reasoning architecture. For instance, with six trajectories, the action generation proportion is just 15.61~\% in multi-reasoning but, it reaches as high as 42.06~\% in single-reasoning. When the number of trajectories is eleven, the contribution remains around 15.82~\% without a visible increase in multi-reasoning, whereas it reaches 52.30~\% in single-reasoning, accounting for more than half of the total inference latency. These results indicate that, in the single-reasoning architecture, the action generation latency can be the dominant latency component as the number of trajectories increases, highlighting the importance of our action generation optimization.

{\bf Action generation optimization results.} Fig.~\ref{fig:diff_overview} shows the latency reduction achieved by our two optimization techniques within a single trajectory generation. By removing unnecessary copy and allocation overhead (denoted as +Static KV Cache), we achieve a 21.66~\% latency reduction compared to the baseline. Furthermore, by additionally applying the CUDA graph capture (denoted as +Graph Capture), the remaining GPU kernel launch overhead is eliminated, yielding a total reduction of 24.65~\%. To provide a more fine-grained analysis, Fig.~\ref{fig:diff_step} shows the latency breakdown across the 10 diffusion iterations. Overall, the per-iteration latency is reduced from 62.83~ms to 49.13~ms with static KV cache management, and further to 47.00~ms with CUDA graph capture. Note that CUDA graph capture requires a preceding warmup iteration to complete memory allocations. Thus, graph capture is performed at the second iteration, resulting in an elevated latency of 69.95~ms as shown in the figure. However, from the third iteration onward, the captured graph is replayed without GPU kernel launch overhead, yielding an average per-iteration latency of 44.83~ms. Thus, despite the recording overhead at the second iteration, CUDA graph capture reduces overall trajectory generation latency. If more diffusion iterations are required for higher-precision trajectory generation, the benefit becomes even more significant.


\begin{figure}
    \centering
    \includegraphics[width=\columnwidth]{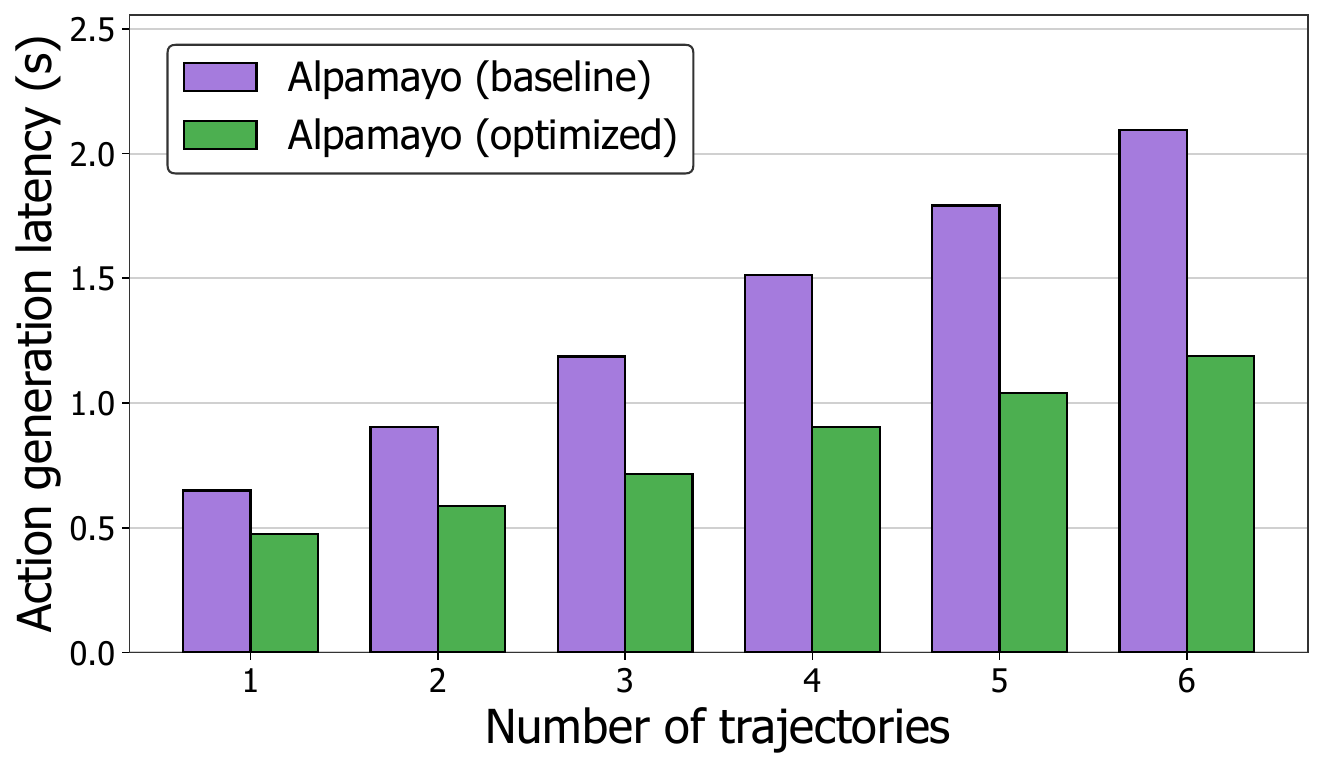}
    \caption{Action generation latency optimization result}
    \label{fig:single_diff}
\end{figure}


We next analyze how our optimization affects the action generation latency with varying numbers of trajectories. Fig.~\ref{fig:single_diff} shows that our optimization consistently reduces the action generation latency.
One interesting observation is that the relative reduction continually grows as the number of trajectories. The reduction proportion is just 27.69~\% (from 0.65~s to 0.47~s) when the number of trajectories is 1. However, it continually increases up to 43.06~\% (from 2.09~s to 1.19~s) at the six-trajectory case. If the trajectory generations are sequential, the latency reduction proportion itself should remain constant, regardless of the number of trajectories. However, Alapmayo performs multi-trajectory diffusion in a batched manner, which significantly increases the memory bandwidth demand as the number of trajectories grows. This leads to a memory bandwidth bottleneck. In such scenarios, our static KV cache management not only reduces latency but also substantially alleviates the memory bandwidth burden. Consequently, the impact of our optimization is further amplified when dealing with a large number of trajectories.



\begin{figure}
    \centering
    \includegraphics[width=\columnwidth]{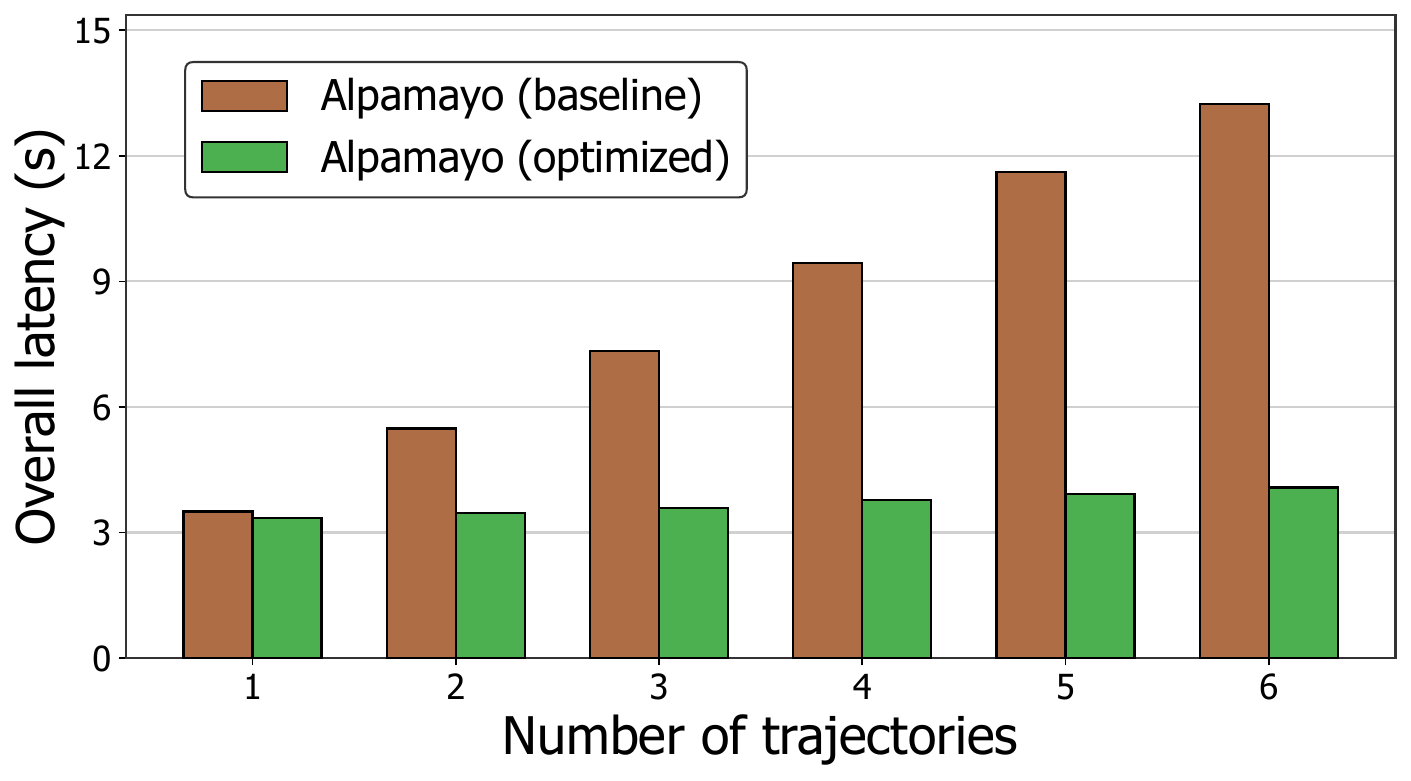}
    \caption{Overall latency optimization result}
    \label{fig:final}
\end{figure}

{\bf Overall optimization result.} Fig.~\ref{fig:final} shows the overall latency comparison between the baseline Alpamayo and our optimized version incorporating both the single-reasoning architecture and the action generation optimization. Our approach consistently reduces latency across all trajectory counts, with the improvement becoming increasingly significant as the number of trajectories grows. In particular, at six trajectories, we achieve up to a 69.23~\% reduction in overall latency (from 13.33~s to 4.10~s) compared to the baseline, demonstrating the effectiveness of combining single-reasoning with efficient action generation. This reduction is critical, as the baseline latency poses a significant barrier to practical deployment, whereas our optimized latency brings VLA-based autonomous driving models closer to real-world applicability.


\section{Related Work}
\label{related}
In this section, we situate our work within reasoning-based E2E autonomous driving systems. Conventional E2E autonomous driving systems lack transparency in their decision-making processes, motivating recent approaches to incorporate VLMs and VLAs to generate human-readable reasoning that enables interpretability of driving decisions ~\cite{Alpamayo, EMMA, DriveGPT4, GPTDriver, DriveVLM, Senna, AutoVLA, AutoDriveR2, AdaThinkDrive, FutureSightDrive, ReasonPlan, ORION, DiffVLA, IRL-VLA, ColaVLA} . Existing approaches can be classified into two categories based on how reasoning relates to trajectory generation: (1) multi-reasoning, where a separate reasoning sequence is generated for each trajectory, and (2) single-reasoning, where a single reasoning sequence is shared across all trajectories.\par

\subsection{Multi-Reasoning E2E Autonomous Driving}
The majority of reasoning-based E2E autonomous driving systems adopt the multi-reasoning paradigm~\cite{Alpamayo, EMMA, DriveGPT4, GPTDriver, DriveVLM, Senna, AutoVLA, AutoDriveR2, AdaThinkDrive, FutureSightDrive, ReasonPlan}. This paradigm is motivated by the intuition that diverse reasoning traces lead to diverse behaviors. However, since all reasoning sequences are conditioned on the same inputs, their variability is primarily driven by stochasticity in token sampling during reasoning generation rather than systematically different scene interpretations. Consequently, increased reasoning diversity may not consistently translate into meaningful behavioral diversity, as we empirically demonstrate in Section~V. Moreover, each trajectory sample requires a separate reasoning pass, resulting in substantial inference overhead as the number of samples grows.\par

\subsection{Single-Reasoning E2E Autonomous Driving}
The limitations of multi-reasoning paradigm motivate the adoption of the single-reasoning paradigm~\cite{ORION, DiffVLA, IRL-VLA, ColaVLA}. In this approach, a single reasoning representation is computed once and shared across all trajectories, with trajectory diversity introduced through stochasticity during trajectory generation. This design reduces inference overhead by eliminating the need to generate multiple reasoning sequences.\par
However, existing approaches adopt this paradigm in a form that reduces interpretability. ORION~\cite{ORION} compresses reasoning into a single planning token that conditions generation of trajectories. DiffVLA~\cite{DiffVLA} conditions its trajectories with VLM-generated driving commands alongside anchor trajectory embeddings. IRL-VLA~\cite{IRL-VLA} encodes semantic commands from a VLM module and geometric features from a BEV encoder into a unified conditioning signal for a diffusion planner. ColaVLA~\cite{ColaVLA} replaces explicit reasoning with latent meta-action representations that guide its planner. In these approaches, the representation that directly conditions trajectory generation is a compressed surrogate derived from reasoning rather than the full internal representation of it. As a result, the human-readable reasoning, where it exists, cannot be treated as a reliable explanation of the generated trajectories, limiting its utility for safety validation.\par
In contrast, we apply the single-reasoning paradigm to Alpamayo without reducing interpretability. We preserve the CoT reasoning of Alpamayo to directly condition trajectory generation, and further reduce inference latency by identifying and eliminating architectural inefficiencies in the diffusion-based trajectory generation process. As a result, our method achieves lower inference latency than the baseline without sacrificing trajectory diversity, prediction quality, or interpretability.\par

\section{Conclusion}
\label{conclusion}
In this paper, we present a comprehensive latency analysis and optimization of Alpamayo, a reasoning-based E2E autonomous driving system. Through detailed profiling of its inference process, we identified two dominant bottlenecks: the multi-reasoning architecture, which causes inference latency to scale linearly with the number of trajectories, and the diffusion-based action generation, which incurs substantial inter-iteration overhead from unnecessary memory copy operations and inefficient GPU kernel launches. To address these bottlenecks, we proposed two targeted optimizations. First, we redesigned Alpamayo into a single-reasoning architecture that generates a shared reasoning once and reuses it across all trajectories, eliminating redundant computation. Second, we eliminated inter-iteration overhead in diffusion-based action generation through memory pre-allocation and CUDA graph capture. Evaluated through both open-loop and closed-loop experiments, the combined optimizations achieve up to a 69.23~\% reduction in end-to-end inference latency while maintaining trajectory diversity and prediction quality. These results demonstrate that jointly analyzing system architecture and runtime execution is essential for improving the practical efficiency of reasoning-based E2E autonomous driving systems. We hope this work will inspire future studies on VLA-based autonomous driving within the real-time systems community.

\balance
\bibliographystyle{IEEEtran}
\bibliography{alpamayo}

\end{document}